\documentclass[journal]{IEEEtran}

\usepackage[cmex10]{amsmath}
\DeclareMathOperator*{\argmin}{arg\,min}
\usepackage{multirow}
\usepackage{mathtools}
\usepackage{amssymb}
\usepackage{hhline}
\usepackage{enumitem}
\usepackage{tabularx}
\usepackage{url}
\usepackage{hyperref}
\usepackage{algpseudocode} 
\usepackage{algorithm}     
\usepackage{float}

\ifCLASSOPTIONcompsoc
  \usepackage[nocompress]{cite}
\else
  \usepackage{cite}
\fi

\ifCLASSINFOpdf
  \usepackage[pdftex]{graphicx}
  \graphicspath{{./figures/}}
  \DeclareGraphicsExtensions{.pdf,.jpeg,.jpg,.png}
\else
  \usepackage[dvips]{graphicx}
  \graphicspath{{./figures/}}
  \DeclareGraphicsExtensions{.eps,.jpeg,.jpg,.png}
\fi

\usepackage{adjustbox} 

\usepackage[usenames, dvipsnames]{color}
\newcommand{\ci}[1]{{\color{Gray}#1}} 

\usepackage{tikz}
\newcommand\copyrighttext{%
  \footnotesize \textcopyright~2018 IEEE. Personal use of this material is permitted.
  Permission from IEEE must be obtained for all other uses, in any current or future
  media, including reprinting/republishing this material for advertising or promotional
  purposes, creating new collective works, for resale or redistribution to servers or
  lists, or reuse of any copyrighted component of this work in other works.
  DOI: \href{https://doi.org/10.1109/TASE.2018.2876430}{10.1109/TASE.2018.2876430}}
\newcommand\copyrightnotice{%
\begin{tikzpicture}[remember picture,overlay]
\node[anchor=south,yshift=6pt] at (current page.south) {\fbox{\parbox{\dimexpr\textwidth-\fboxsep-\fboxrule\relax}{\copyrighttext}}};
\end{tikzpicture}%
}

\begin{document}

\title{Automatic composition and optimisation of multicomponent predictive systems \\with an extended Auto-WEKA}

\author{Manuel~Martin~Salvador,
        Marcin~Budka,
        and~Bogdan~Gabrys
\thanks{M. Martin Salvador is with Blink Technologies, Palo Alto, CA, USA. (email: manuel@blinkeye.ai)

M. Budka is with Bournemouth University, Poole, United Kingdom. (email: mbudka@bournemouth.ac.uk)

B. Gabrys is with Advanced Analytics Institute, University of Technology Sydney, Australia. (email: Bogdan.Gabrys@uts.edu.au)}
}

\markboth{IEEE Transactions on Automation Science and Engineering}%
{Martin Salvador \MakeLowercase{\textit{et al.}}: Automatic composition and optimisation of multicomponent predictive systems}

\renewcommand{\topfraction}{.75}

\maketitle

\copyrightnotice

\begin{abstract}
Composition and parametrisation of multicomponent predictive systems (MCPSs) consisting of chains of data transformation steps is a challenging task. Auto-WEKA is a tool to automate the Combined Algorithm Selection and Hyperparameter (CASH) optimisation problem. In this paper we extend the CASH problem and Auto-WEKA to support MCPS including preprocessing steps for both classification and regression tasks. We define the optimisation problem in which the search space consists of suitably parametrised Petri nets forming the sought MCPS solutions.
In the experimental analysis we focus on examining the impact of considerably extending the search space (from approximately 22,000 to 812 billion possible combinations of methods and categorical hyperparameters). In a range of extensive experiments three different optimisation strategies are used to automatically compose MCPSs for 21 publicly available datasets. The diversity of the composed MCPSs found is an indication that fully and automatically exploiting different combinations of data cleaning and preprocessing techniques is possible and highly beneficial for different predictive models. We also present the results on 7 datasets from real chemical production processes. Our findings can have a major impact on development of high quality predictive models as well as their maintenance and scalability aspects needed in modern applications and deployment scenarios.
\end{abstract}

 
\renewcommand{\abstractname}{Note to Practitioners}
\begin{abstract}
The extension of Auto-WEKA to compose and optimise MCPSs developed as part of this paper is freely available on GitHub under GPL licence and we encourage practitioners to use it on a broad variety of classification and regression problems. The software can either be used as a blackbox -- where search space is made of all possible WEKA filters, predictors and meta-predictors (e.g. ensembles) -- or as an optimisation tool on a subset of pre-selected machine learning methods. The application has a graphical user interface, but also can run from command line and can be embedded in any project as a Java library. There are three main outputs once an Auto-WEKA run has finished: a) the trained MCPS ready to make predictions on unseen data; b) the WEKA configuration (i.e. parametrised components); c) the Petri net in a PNML (Petri Net Markup Language) format which can be analysed using any tool supporting this standard language. There are however some practical considerations affecting the quality of the results that must be taken into consideration such as the CPU time budget or the search starting point. These are extensively discussed in the paper.
\end{abstract}


\IEEEpeerreviewmaketitle

\section{Introduction}

\IEEEPARstart{P}{erformance} of data-driven predictive models heavily relies on the quality and quantity of data used to build them. In real applications, even if data is abundant, it is also often imperfect and considerable effort needs to be invested into a labour-intensive task of cleaning and preprocessing such data in preparation for subsequent modelling. A survey\footnote{\url{http://www.kdnuggets.com/polls/2003/data_preparation.htm}} of data mining practitioners carried out in 2003 indicates that preprocessing tasks can account for as much as 60-80\% of the total time spent on developing a predictive model. More recent surveys from 2012~\cite{Munson2012} and 2016\footnote{`Cleaning Big Data: Most Time-Consuming, Least Enjoyable Data Science Task, Survey Says' by Gil Press. Forbes 2016. \url{http://bit.ly/forbes-data-preparation}} confirm these numbers. The practitioners also note that preprocessing is the least enjoyable part of data science. The reason for this being such a lengthy process is all the manual work necessary to identify the defects in the raw data and look for the best solutions to approach them. Despite 13 years that passed between the surveys, no significant advances have been made to address this issue. Therefore, it is desirable to automate as many of the tasks of data preprocessing as possible in order to reduce the human involvement and the level of necessary interactions. The consequence of this would be speeding up of the data mining process and making the procedures more robust.

After the data has been preprocessed in an appropriate way, the next step in a data mining process is modelling (i.e. finding an appropriate classifier or regressor). Similarly to preprocessing, this step can also be very labour-intensive, requiring evaluation of multiple alternative models. Hence automatic model selection has been attempted in different ways, for example using active testing~\cite{Leite2012}, meta-learning~\cite{Lemke2010}, information theory~\cite{MacQuarrie1998} or following a multi-criteria decision making process~\cite{Ali2017}. More recently, Google has launched a new service called Cloud AutoML for automatically building deep neural networks for image classification problems, using transfer learning~\cite{Zoph2017}. There is then an increasing attention from academy and industry in this topic. We have observed that a common theme in the literature is comparison of different models using data always preprocessed in the same way. However, some models may perform better if they are built using data specifically preprocessed with a particular model type in mind. In addition, hyperparameters play an important role in most of the models, and setting them manually is time-consuming mainly for two reasons: (1)~there are typically multiple hyperparameters which can take many values (with an extreme case being continuous hyperparameters), and (2)~they are validated using cross-validation (CV).

In many scenarios one needs to sequentially apply multiple preprocessing methods to the data (e.g. outlier detection $\rightarrow$ missing value imputation $\rightarrow$ dimensionality reduction), effectively forming a preprocessing chain. Data-driven workflows have been used to guide data processing in a variety of fields. Some examples are astronomy \cite{Berriman2007}, biology \cite{Shade2015}, clinical research \cite{Teichmann2010}, archive scanning \cite{Messaoud2011}, telecommunications \cite{Maedche2000}, banking \cite{Wei2013} and process industry \cite{Budka2014}, to name a few. The common methodology in all these fields consists of following a number of steps to prepare a dataset for data mining. In the field of predictive modelling, the workflow resulting from connecting different methods is known as a Multi-Component Predictive System (MCPS) \cite{Tsakonas2012,MartinSalvador2017}. At the moment, tools like WEKA\footnote{\url{http://weka.sourceforge.net}}, RapidMiner\footnote{\url{https://rapidminer.com}} or Knime\footnote{\url{https://www.knime.org}} allow to create and run MCPSs including a large variety of operators.

The motivation for automating composition of MCPS is twofold. In the first instance, it will help to reduce the amount of time spent on the most labour-intensive activities related to predictive modelling, and therefore allow to dedicate human expertise to other tasks. The second motivation is to achieve better results than a human expert could, given a limited amount of time. The number of possible methods and hyperparameter combinations increases exponentially with the number of components in an MCPS and in majority of cases it is not computationally feasible to evaluate all of them. Therefore it makes sense to approach preprocessing, model selection and hyperparameter optimisation problems jointly.

The Combined Algorithm Selection and Hyperparameter optimisation (CASH) problem presented in~\cite{Thornton2013} consists of finding the best combination of learning algorithm $A^*$ and hyperparameters $\lambda^*$ that optimise an objective function (e.g. Eq.~\ref{eq:cash} minimises the $k$-fold cross-validation error) for a given dataset $\mathcal{D}$. Formally, CASH problem is given by
\begin{equation}
\label{eq:cash}
A^*_{\lambda^*} = \argmin_{A^{(j)} \in \mathcal{A}, \lambda \in \Lambda^{(j)}} \frac{1}{k} \sum_{i=1}^{k} \mathcal{L}(A_{\lambda}^{(j)}, \mathcal{D}_{train}^{(i)}, \mathcal{D}_{valid}^{(i)})
\end{equation}
where $\mathcal{A} = \{A^{(1)}, \dotsc, A^{(k)}\}$ is a set of algorithms with associated hyperparameter spaces $\Lambda^{(1)}, \dotsc, \Lambda^{(k)}$. The loss function $\mathcal{L}$ takes as arguments an algorithm configuration $A_{\lambda}$ (i.e. an instance of a learning algorithm and hyperparameters), a training set $\mathcal{D}_{train}$ and a validation set $\mathcal{D}_{valid}$.

In this paper we extend the CASH problem to support MCPSs, i.e. joint optimisation of predictive models (classifiers and regressors) and preprocessing chains. We embrace the representation of MCPS as Petri nets which we proposed in \cite{MartinSalvador2017} and thus
\begin{equation}
  \theta = (P, T_{\boldsymbol{\lambda}}, F)
\end{equation}
 \
\begin{figure}
\centering
\includegraphics[width=0.6\linewidth]{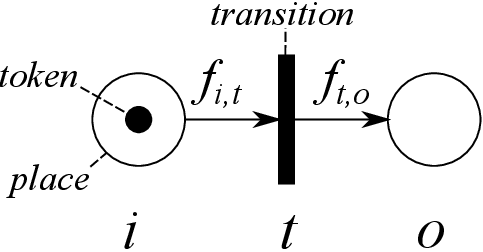}
\caption{Multicomponent predictive system with a single transition}
\label{fig:MCPS-simple-net}
\end{figure}

In our previous work~\cite{MartinSalvador2016} we presented the first approach for hyperparameter optimisation of WEKA classifiers that modify their inner data preprocessing behaviour by recursively expanding the search space constructed by Auto-WEKA (a tool for solving the CASH problem defined by WEKA algorithms and hyperparameters). In this paper we present a further development of Auto-WEKA to support any combination of preprocessing methods (known as WEKA filters). This leads to significantly enlarging the search space of the CASH problem and automating the composition and optimisation of such complex MCPSs.

The paper is organised as follows. The next section reviews previous work in automating the CASH problem and highlights the available software. Section~\ref{sec:MCPS-composition} extends CASH problem to MCPSs and describes the challenges related to automation of their composition. In Section~\ref{sec:autoweka}, our contributions to Auto-WEKA software, now allowing to create and optimise arbitrary chains of preprocessing steps followed by a predictive model, are presented.
The methodology used to automate MCPS composition is discussed in Section~\ref{sec:methodology} followed by the results of extensive experimental analysis in Section~\ref{sec:results}. In Section~\ref{sec:process-industry}, we present the results of applying this approach to real datasets from the process industry. Finally, the paper concludes in Section~\ref{sec:conclusion}.

\section{Bayesian optimisation strategies}
\label{sec:related-work}

The CASH problem as shown in Eq.~\ref{eq:cash} can be approached in different ways. One example is a grid search, i.e. an exhaustive search over all the possible combinations of discretized parameters. Such technique can however be computationally prohibitive in large search spaces or with big datasets. A simpler mechanism like random search, where the search space is randomly explored in a limited amount of time, has been shown to be more effective in high-dimensional spaces~\cite{Bergstra2012}.

A promising approach gaining popularity in the recent years is based on a Bayesian optimization framework~\cite{Brochu2010, Shahriari2016}. This approach -- outlined in Algorithm~\ref{alg:bayesian-optimisation} -- aims to find
\begin{equation}
\theta^* = \operatornamewithlimits{argmin}\limits_{\theta \in \Theta} \mathcal{L}(\theta, \mathcal{D}_{train}, \mathcal{D}_{test})
\end{equation}
that globally minimises the loss function $\mathcal{L}$. It assumes that the posterior distribution $p(\mathcal{L} \mid R_{1:n})$ can be estimated by the likelihood function $p(R_{1:n} \mid \mathcal{L})$ and the prior distribution $p(\mathcal{L})$ using Bayes' theorem
\begin{equation}
p(\mathcal{L} \mid R_{1:n}) \propto p(R_{1:n} \mid \mathcal{L}) p(\mathcal{L})
\end{equation}
where $R_{1:n} = \{(\theta_1, c_{\theta_1}), ..., (\theta_n, c_{\theta_n}) \}$ is the set of run configurations and its associated costs. Since evaluating the loss function is costly, an acquisition function $\alpha_{p(\mathcal{L})} : \Theta \rightarrow \mathbb{R}$ quantifying the utility of an evaluation is used instead as a cheaper alternative. This function has an inherent trade-off between exploration (where there is more uncertainty) and exploitation (where the cost is expected to be low). There are different types of acquisition functions based on the likelihood of improvement~\cite{Kushner1964}, the upper confidence bound criterion~\cite{Lai1985}, or information gain~\cite{Hennig2012}.

\begin{algorithm}
\begin{algorithmic}[1]
\caption{Bayesian optimisation}
\label{alg:bayesian-optimisation}
\For{$n = 1,2,...$}
\State $\theta_{n+1} = \operatornamewithlimits{argmax}\limits_{\theta \in \Theta} \alpha(\theta)$ \Comment select most promising configuration
\State $c_{\theta_{n+1}} = \mathcal{L}(\theta_{n+1}, \mathcal{D}_{train}, \mathcal{D}_{test})$ \Comment compute cost
\State $R_{n+1} = \{R_n, (\theta_{n+1}, c_{\theta_{n+1}})\}$ \Comment update list of run configurations
\State update $p(\mathcal{L} \mid R_{1:n+1})$
\EndFor
\end{algorithmic}
\end{algorithm}

In particular, Sequential Model-Based Optimization (SMBO)~\cite{Hutter2011} is a Bayesian optimisation framework that incrementally builds a regression model $\psi$  -- known as surrogate model -- using instances from $R$. Then, such model is used to predict the performance of promising candidate configurations. The selection of promising configurations is guided by an acquisition function $\alpha_{\psi} : \Theta \rightarrow \mathbb{R}$. A function that has been shown to work well in SMBO framework~\cite{Hutter2009} is the expected improvement (EI)~\cite{Mockus1978} given by
\begin{equation}
\label{eq:expected-improvement}
\alpha_{\psi}(\theta \mid R_n) = \mathbb{E}_{\psi}[I(\theta)] = \mathbb{E}_{\psi}[\max\{0, c_{min} - c_{\theta}\}]
\end{equation}
where $c_{min}$ is the cost of the best configuration found so far. The advantage of this function is that it can be evaluated without computing the loss function for each $\theta$ (i.e. running the configuration) since $c_{\theta}$ can be estimated using $\psi$. A common technique to select the next promising configuration consists of evaluating EI for thousands of random samples and then returning the best one~\cite{Hutter2011}. Algorithm~\ref{alg:smbo} shows the SMBO procedure that returns the best configuration found $\theta_{min}$ (also known as `incumbent').

\begin{algorithm}
\begin{algorithmic}[1]
\caption{Sequential Model-Based Optimisation}
\label{alg:smbo}
\State $\theta_{min}$ = initial configuration (usually random sample from $\Theta$)
\State $R = \{ [\theta_{min}, \mathcal{L}(\theta_{min}, \mathcal{D})] \}$ \Comment initialise set of run configurations and associated costs
\Repeat
\State $\psi = FitModel(R)$
\State $\theta = FindNextConfiguration(\alpha, \psi, \theta_{min}, \Theta)$
\State $c_{\theta} = \mathcal{L}(\theta, \mathcal{D})$ \Comment compute cost
\State $R.add([\theta, c_{\theta}])$ \Comment update list of run configurations
\State $\theta_{min} = \operatornamewithlimits{argmin}\limits_{\theta} c_{\theta} \mid [\theta, c_{\theta}] \in R$ \Comment update best configuration found
\Until{budget exhausted}
\State
\Return $\theta_{min}$
\end{algorithmic}
\end{algorithm}

Some hyperparameters influence the optimisation problem conditionally, i.e. only when some other hyperparameters take certain values.
For example Gaussian kernel width in Support Vector Machine (SVM) is only relevant if SVM is using Gaussian kernels in the first place. Search spaces containing this type of hyperparameters are known as conditional spaces~\cite{Shahriari2016}.

The ability of SMBO methods to work in conditional spaces is given by the surrogate model they use:  models like Random Forests or the Tree Parzen Estimator (TPE) support conditional attributes. A successful SMBO approach using random forests is SMAC (Sequential Model-based Algorithm Configuration by \cite{Hutter2011}) where an ensemble of decision trees makes it possible to model conditional variables. Another state-of-the-art approach uses TPE~\cite{Bergstra2011}, where a graph-structured model matches the conditional structure of the search space. Other SMBO approaches use Gaussian processes as surrogate models (e.g. \cite{Snoek2012}). However, they cannot work in conditional spaces because standard kernels are not defined over variable-length spaces~\cite{Shahriari2016} and therefore are not used in this paper.

Currently available software tools supporting SMBO methods are listed in Table~\ref{tab:available-software}. It should be noted however, that to the best of our knowledge, there are no comprehensive studies or tools\footnote{At the moment of writing this paper there were none, but some new tools have appeared in the meantime (see e.g. TPOT \cite{Olson2016} \url{https://github.com/EpistasisLab/tpot} and DataRobot \url{https://www.datarobot.com}).} which would tackle the problem of flexible composition of many data preprocessing steps (e.g.~data cleaning, feature selection, data transformation) and their simultaneous parametric optimisation, which is one of the key issues addressed in this paper.

\begin{table*}
\centering
\caption{Popular open-source tools supporting SMBO methods}
\begin{adjustbox}{max width=\textwidth}
\begin{tabular}{l l l l}
\hline
\textbf{Name} & \textbf{Surrogate model} & \textbf{Language} & \textbf{URL} \\ \hline
SMAC & Random forest & Java & \url{http://www.cs.ubc.ca/labs/beta/Projects/SMAC} \\
Hyperopt & Tree Parzen estimator & Python & \url{https://github.com/hyperopt/hyperopt} \\
Spearmint & Gaussian process & Python & \url{https://github.com/HIPS/Spearmint} \\
Bayesopt & Gaussian process & C++ & \url{https://bitbucket.org/rmcantin/bayesopt} \\
PyBO & Gaussian process & Python & \url{https://github.com/mwhoffman/pybo} \\
MOE & Gaussian process & Python / C++ & \url{https://github.com/Yelp/MOE} \\
Scikit-Optimize & Various & Python & \url{https://scikit-optimize.github.io} \\
Auto-WEKA* & SMAC,TPE & Java & \url{https://github.com/automl/autoweka} \\
Auto-Sklearn* & SMAC & Python & \url{https://github.com/automl/auto-sklearn} \\
\hline
\multicolumn{4}{l}{* Toolkits for automating algorithm selection in WEKA and Scikit-learn, respectively.}
\end{tabular}
\end{adjustbox}
\label{tab:available-software}
\end{table*}

\section{Automating MCPS composition}
\label{sec:MCPS-composition}

Building an MCPS is typically an iterative, labour and knowledge intensive process. Despite a substantial body of research in the area of automated and assisted MCPS creation and optimisation (see e.g.~\cite{Serban2013} for a survey), a reliable fully automated approach still does not exist.

To accommodate the definition of MCPS into a CASH problem we generalise $\mathcal{A}$ from Eq.~\ref{eq:cash} to be a set of MCPSs $\Theta = \{\theta^{(1)}, \theta^{(2)}, ...\}$ rather than individual algorithms. Hence each MCPS $\theta^{(j)} = (P, T_{\boldsymbol{\lambda}}, F)^{(j)}$ has now a hyperparameter space $\Lambda^{(j)}$, which is a concatenation of the hyperparameter spaces of all its transitions $T$. The CASH problem is now concerned with finding $(P, T_{\boldsymbol{\lambda}^*}, F)^*$ such as:

\begin{equation}
\label{eq:MCPS-CASH}
\argmin_{(P, T, F)^{(j)} \in \Theta, \boldsymbol{\lambda} \in \Lambda^{(j)}} \frac{1}{k} \sum_{i=1}^{k} \mathcal{L}((P, T_{\boldsymbol{\lambda}}, F)^{(j)}, \mathcal{D}_{train}^{(i)}, \mathcal{D}_{valid}^{(i)})
\end{equation}

\begin{figure}
\centering
\includegraphics[width=0.5\linewidth]{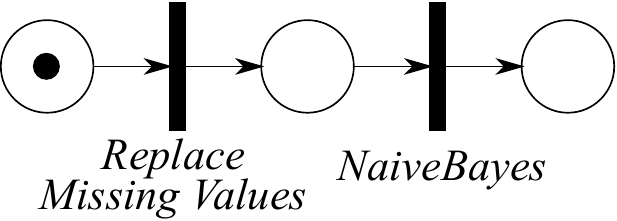}
\caption{Example of a simple MCPS}
\label{fig:MCPS-simple}
\end{figure}

\begin{figure}
\centering
\includegraphics[width=\linewidth]{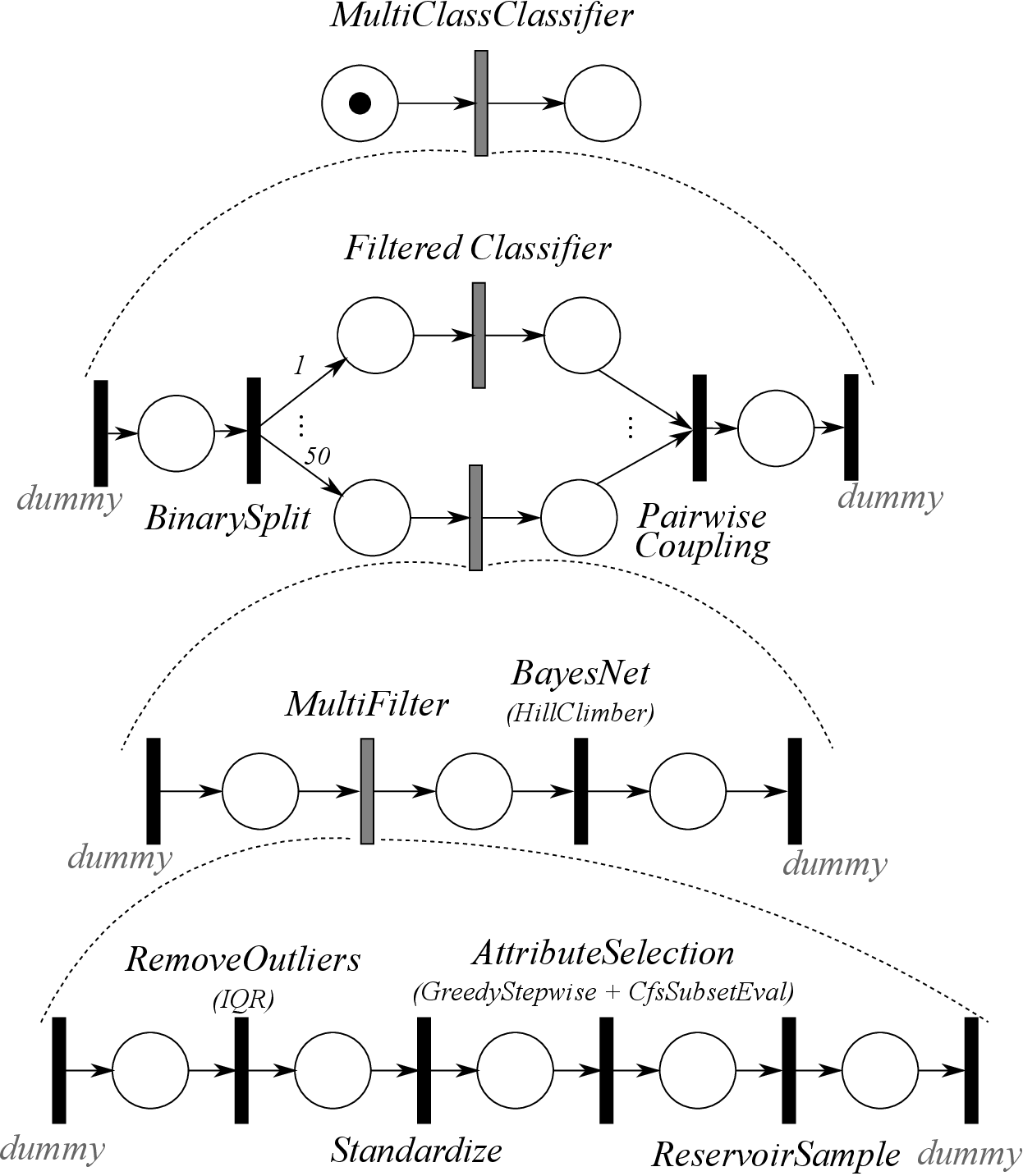}
\caption{Example of a complex MCPS}
\label{fig:MCPS-kddcup09app-FULL}
\end{figure}

The main reason for MCPS composition being a challenging problem is the computational power and time needed to explore high dimensional search spaces. To begin with, an undetermined number of components can make the workflow very simple (see Figure~\ref{fig:MCPS-simple}) or very complex (see Figure~\ref{fig:MCPS-kddcup09app-FULL}). Secondly, the order in which the nodes should be connected is unknown a priori. Also, even transitions belonging to the same category (e.g. missing value imputation) can vary widely in terms of the number and type of hyperparameters (e.g. continuous, categorical or conditional), with defining of a viable range for each of the hyperparameters being an additional problem in itself. This complexity makes techniques like grid search not feasible. Even `intelligent' strategies can struggle with exploration because the high dimensional search space is likely to be plagued with a multitude of bad local minima.

The size of the search space (i.e. $|\Theta|$) can be reduced by applying a range of constraints like limiting the number of components, restricting the list of methods using meta-learning~\cite{Feurer2014}, prior knowledge~\cite{Swersky2013} or surrogates (i.e. cheap-to-evaluate models~\cite{Eggensperger2012}). However, this study investigates the impact of extending the search space, not by including more predictive models, but considering preprocessing methods instead. Nonetheless, some constrains are applied like limiting the number and order of components which will be explained in Section~\ref{sec:methodology}.

We use the predictive performance as a sole optimisation objective as shown in Eq.~\ref{eq:MCPS-CASH}, noting however, that some problems may require to optimise several objectives at the same time (e.g. error rate, model complexity and runtime~\cite{Al-Jubouri2014}). In this paper we use our extended Auto-WEKA version -- described in the next section -- which supports automatic composition and optimisation of MCPSs with WEKA filters and predictive models as components.

\begin{figure}
\centering
\includegraphics[width=0.9\linewidth]{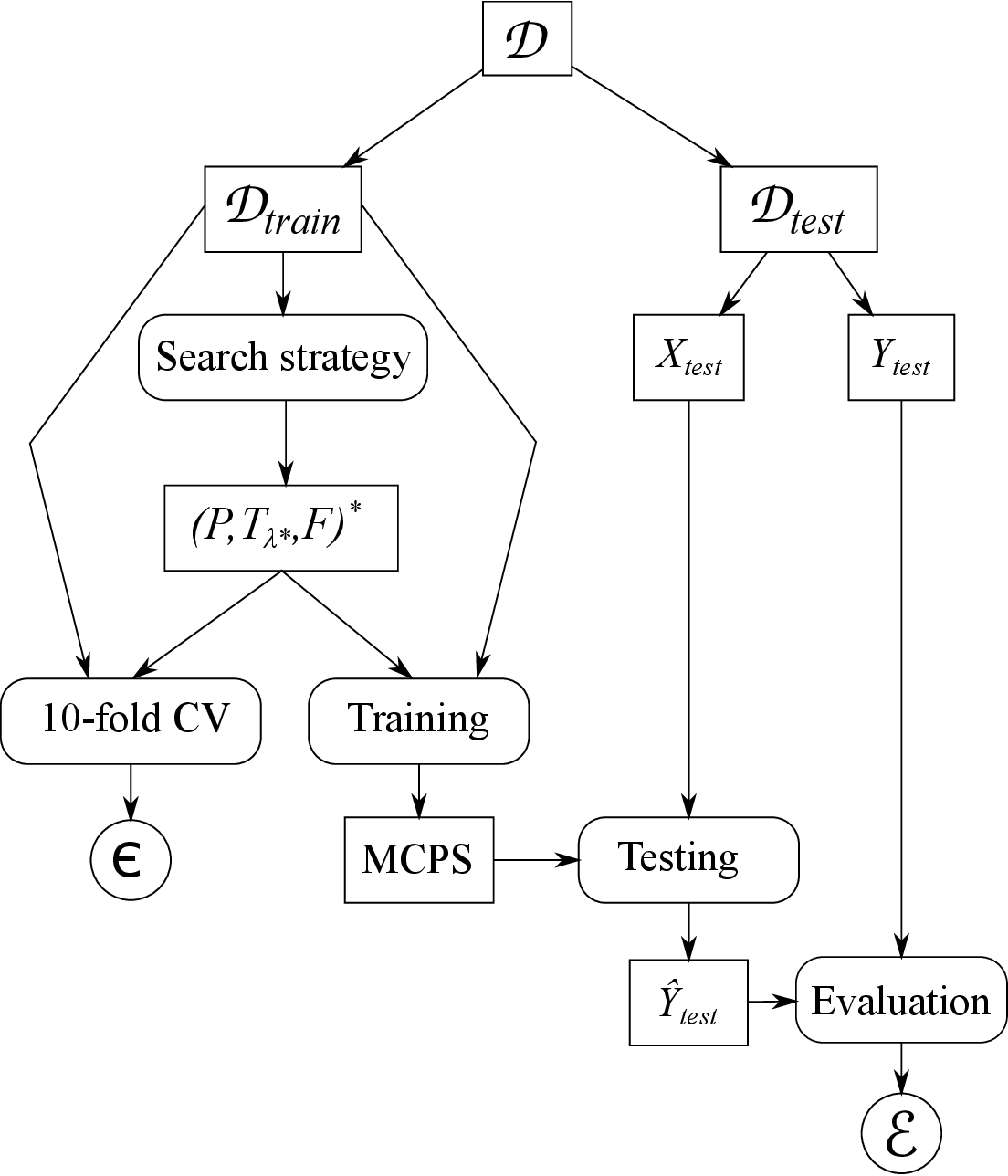}
\caption{MCPS training and testing process}
\label{fig:process}
\end{figure}

Once the MCPS is composed and its hyperparameters optimised, it is trained with a set of labelled instances. Then the MCPS is ready to make predictions as shown in Figure~\ref{fig:process}.

\subsection{Extension and generalisation of Auto-WEKA}
\label{sec:autoweka}

Auto-WEKA is a software developed by Thornton et al.~\cite{Thornton2013} which allows algorithm selection and hyperparameter optimisation both in regression and classification problems.

In this work we have extended Auto-WEKA to support MCPSs. This leads to an increasing of the search space from 21,560 up to 812 billion possible solutions as seen in Table~\ref{tab:search-spaces} and further discussed in Section~\ref{sec:methodology}. This work has been developed taking as base Auto-WEKA version 0.5. Independently and in parallel to our work, there was further development of Auto-WEKA which current version is now 2.6~{\cite{Kotthoff2016}. Nonetheless, this new version does not affect the work carried out in this paper as its main novelty is the integration with WEKA user interface.

Both versions provide a one-click solution for automating algorithm selection and hyperparameter optimisation. However, version 0.5 is much more flexible, offering as well multiple customisations possibilities like preselection of WEKA predictors, choosing the optimisation strategy or setting the optimisation criteria. Auto-WEKA 0.5 also supports various usage scenarios depending on user knowledge, needs and available computational budget. One can for example, run several optimisations in parallel, ending up with multiple solutions that can then be analysed individually or used to build an ensemble~\cite{Feurer2015b}.

Our new extensions now allow any WEKA filter to be included as part of the composition process. In addition, we have developed a new WEKA filter that creates a flexible chain of common preprocessing steps including missing values handling, outlier detection and removal, data transformation, dimensionality reduction and sampling.

The following external WEKA packages\footnote{\url{http://weka.sourceforge.net/packageMetaData/}} have been included as part of the developed Auto-WEKA extensions to increase the number of preprocessing methods: \emph{EMImputation}, \emph{RBFNetwork}, \emph{StudentFilters}, \emph{baggedLocalOutlierFactor}, \emph{localOutlierFactor}, \emph{partialLeastSquares} and \emph{wavelet}. Furthermore, we have developed two new WEKA filters: (i) an outlier detection and removal filter in a single step; and (ii) a sampling filter in which instances are periodically selected given a fixed interval of time. These new filters are common operations in the preprocessing of datasets from the process industry.

Moreover, our extension generates an MCPS in a PNML (Petri Net Markup Language) format which can be analysed using any tool supporting this standard language (e.g. WoPeD\footnote{\url{http://woped.dhbw-karlsruhe.de/woped/}}).

Therefore, there are three main outputs once a new, extended Auto-WEKA run has finished: a) the trained MCPS ready to make predictions on unseen data; b) WEKA configuration (i.e. parametrised components); c) the Petri net in a PNML format.

While the space restriction does not allow us to include more implementation details, the source code and all the scripts for the analysis of the extended Auto-WEKA results such as the creation of plots and tables have been released in our repository\footnote{\url{https://github.com/dsibournemouth/autoweka}}.


\section{Methodology}
\label{sec:methodology}

The purpose of this experimental study is to analyse the feasibility of SMBO strategies for solving the MCPS related CASH problem as defined in Eq.~\ref{eq:MCPS-CASH} and the quality of the solutions found using the proposed extended Auto-WEKA with all its features supporting the MCPSs composition and parametric/hyperparametric optimisation of the workflows.

The three main characteristics which define a CASH problem are: a)~the search space, b)~the objective function and c)~the optimisation algorithm. In this study we have considered three search spaces of very different sizes (see Table~\ref{tab:search-spaces}):
\begin{itemize}[leftmargin=16px]
\item PREV: This is the search space used in~\cite{Thornton2013} where predictors and meta-predictors (which take outputs from one or more base predictive models as their input) were considered (756 hyperparameters). It can also include the best feature selection method found after running `AttributeSelection' WEKA filter (30 hyperparameters) for 15 minutes before the optimisation process begins. We use it as a baseline.
\item NEW: This search space only includes predictors and meta-predictors. In contrast with PREV space, no previous feature selection stage is performed. We would like to note however that some WEKA classifiers perform internal preprocessing steps as we showed in our previous work~\cite{MartinSalvador2016} ((e.g. MultiLayerPerceptron (MLP) removes instances with missing values and scales the attributes to a range [-1,1])). We take into account that a categorical hyperparameter can be either simple or complex (i.e.~when it contains WEKA classes). In the latter case, we increase the search space by adding recursively the hyperparameters of each method belonging to such complex parameter (e.g.~the `DecisionTable' predictor contains a complex hyperparameter whose values are three different types of search methods with further hyperparameters -- see Table~\ref{tab:search-space-predictors} for details). That extension increases the search space to 1186 hyperparameters.
\item FULL: This search space has been defined to support a flow with up to five preprocessing steps, a predictive model and a meta-predictor (1564 hyperparameters). The nodes are connected in the following order: missing value handling $\rightarrow$ outlier detection and handling\footnote{Outliers are handled in a different way than missing values} $\rightarrow$ data transformation $\rightarrow$ dimensionality reduction $\rightarrow$ sampling $\rightarrow$ predictor $\rightarrow$ meta-predictor. This flow is based on our experience with process industry~\cite{Budka2014}, but these preprocessing steps are also common in other fields. If the meta-predictor is either `Stacking' or `Vote', its number of inputs can vary from 1 to 5.
\end{itemize}

\begin{table}
\centering
\caption{Summary of search spaces. MV = Missing Value replacement, OU = Outlier Detection and Removal, TR = Transformation, DR = Dimensionality Reduction, SA = Sampling, P = Predictor, MP = Meta-Predictor, CH = Expanding Complex Hyperparameters. * Size considering only methods, categorical and complex hyperparameters.}
\begin{adjustbox}{max width=\columnwidth}
\begin{tabular}{l | c c c c c c c c r}
\hline
~ & MV & OU & TR & DR & SA & P & MP & CH & Size* \\ \hline
PREV &  ~ & ~ & ~ & $\checkmark$ & ~ & $\checkmark$ & $\checkmark$ & ~ & 21,560 \\
NEW &  ~ & ~ & ~ & ~ & ~ & $\checkmark$ & $\checkmark$ & $\checkmark$ & 2,369,598 \\
FULL & $\checkmark$ & $\checkmark$ & $\checkmark$ & $\checkmark$ & $\checkmark$ & $\checkmark$ & $\checkmark$ & $\checkmark$ & 812 billion \\ \hline
\end{tabular}
\end{adjustbox}
\label{tab:search-spaces}
\end{table}

The methods that can be included in each component are listed in Tables~\ref{tab:search-space-preprocessing} and~\ref{tab:search-space-predictors}. Note that the FULL search space is more than twice as large as the one presented in~\cite{Thornton2013} in terms of the raw number of hyperparameters.

\begin{table}[!htbp]
\centering
\caption{Number of parameters of the available preprocessing methods.}
\begin{tabular}{l c c c}
\hline

\multirow{2}{*}{Method} & Num. & \multicolumn{2}{c}{Categorical} \\ \cline{2-4}

~ & ~ & Simple & Complex \\ \hline

\textbf{Missing values (MV)} & ~ & ~ & ~\\
No handling & 0 & 0 & 0 \\ 
ReplaceMissingValues & 0 & 0 & 0 \\ 
CustomReplaceMissingValues & 0 & M & 0 \\ 
$\hookrightarrow$ (M) Zero & 0 & 0 & 0 \\
$\hookrightarrow$ (M) Mean & 0 & 0 & 0 \\
$\hookrightarrow$ (M) Median & 0 & 0 & 0 \\
$\hookrightarrow$ (M) Min & 0 & 0 & 0 \\
$\hookrightarrow$ (M) Max & 0 & 0 & 0 \\
$\hookrightarrow$ (M) LastKnown & 0 & 0 & 0 \\
EMImputation & 3 & 1 & 0 \\ \hline

\textbf{Outliers (OU)} & ~ & ~ & ~\\
No handling & 0 & 0 & 0 \\ 
RemoveOutliers & 0 & 0 & O \\
$\hookrightarrow$ (O) InterquartileRange (IQR) & 2 & 0 & 0 \\
$\hookrightarrow$ (O) BaggedLOF & 1 & 1 & 0 \\ 
\hline

\textbf{Transformation (TR)} & ~ & ~ & ~\\
No transformation & 0 & 0 & 0 \\ 
Center & 0 & 0 & 0 \\ 
Standardize & 0 & 0 & 0 \\ 
Normalize & 2 & 0 & 0 \\ 
Wavelet & 0 & 0 & 0 \\ 
IndependentComponents & 3 & 1 & 0 \\ \hline

\textbf{Dimensionality reduction (DR)} & ~ & ~ & ~\\
No reduction & 0 & 0 & 0 \\ 
PrincipalComponents (PCA) & 3 & 1 & 0 \\
RandomSubset & 2 & 0 & 0 \\ 
AttributeSelection & 0 & 0 & S,E \\ 
$\hookrightarrow$ (S) BestFirst & 1 & 1 & 0 \\
$\hookrightarrow$ (S) GreedyStepwise & 2 & 3 & 0 \\
$\hookrightarrow$ (S) Ranker & 1 & 0 & 0 \\
$\hookrightarrow$ (E) CfsSubsetEval & 0 & 2 & 0 \\
$\hookrightarrow$ (E) CorrelationAttributeEval & 0 & 0 & 0 \\
$\hookrightarrow$ (E) GainRatioAttributeEval & 0 & 0 & 0 \\
$\hookrightarrow$ (E) InfoGainAttributeEval & 0 & 2 & 0 \\
$\hookrightarrow$ (E) OneRAttributeEval & 2 & 1 & 0 \\
$\hookrightarrow$ (E) PrincipalComponents & 2 & 3 & 0 \\
$\hookrightarrow$ (E) ReliefFAttributeEval & 2 & 1 & 0 \\
$\hookrightarrow$ (E) Sym.UncertAttributeEval & 0 & 1 & 0 \\
$\hookrightarrow$ (E) WrapperSubsetEval & 0 & 0 & 0 \\
PLSFilter & 1 & 4 & 0 \\ \hline

\textbf{Sampling (SA)} & ~ & ~ & ~\\
No sampling & 0 & 0 & 0 \\ 
Resample & 2 & 0 & 0 \\ 
ReservoirSample & 2 & 0 & 0 \\ 
Periodic sampling & 1 & 0 & 0 \\ \hline
\end{tabular}
\label{tab:search-space-preprocessing}
\end{table}

\begin{table}[!tbp]
\centering
\caption{Number of parameters of the available predictors.}
\begin{tabular}{l c c c}
\hline
\multirow{2}{*}{Method} & Num. & \multicolumn{2}{c}{Categorical} \\ \cline{2-4}

~ & ~ & Simple & Complex \\ \hline

\textbf{Predictors (P)} & ~ & ~ & ~\\
BayesNet & 0 & 1 & Q \\ 
$\hookrightarrow$ (Q) local.K2 & 1 & 4 & 0 \\
$\hookrightarrow$ (Q) local.HillClimber & 1 & 4 & 0 \\
$\hookrightarrow$ (Q) local.LAGDHillClimber & 3 & 4 & 0 \\
$\hookrightarrow$ (Q) local.SimulatedAnnealing & 3 & 2 & 0 \\
$\hookrightarrow$ (Q) local.TabuSearch & 3 & 4 & 0 \\
$\hookrightarrow$ (Q) local.TAN & 0 & 2 & 0 \\
NaiveBayes & 0 & 2 & 0 \\ 
NaiveBayesMultinomial & 0 & 0 & 0 \\ 
Logistic & 1 & 0 & 0 \\ 
MLP & 6 & 5 & 0 \\ 
SMO & 1 & 2 & K \\ 
$\hookrightarrow$ (K) NormalizedPolyKernel & 1 & 1 & 0 \\
$\hookrightarrow$ (K) PolyKernel & 1 & 1 & 0 \\
$\hookrightarrow$ (K) Puk & 2 & 0 & 0 \\
$\hookrightarrow$ (K) RBFKernel & 1 & 0 & 0 \\
SimpleLogistic & 0 & 3 & 0 \\ 
IBk & 1 & 4 & A \\ 
$\hookrightarrow$ (A) BallTree & 0 & 1 & 2 \\
$\hookrightarrow$ (A) CoverTree & 1 & 1 & 1 \\
$\hookrightarrow$ (A) KDTree & 2 & 1 & 2 \\
$\hookrightarrow$ (A) LinearNNSearch & 0 & 1 & 0 \\
KStar & 1 & 2 & 0 \\ 
DecisionTable & 0 & 3 & S \\ 
$\hookrightarrow$ (S) BestFirst & 1 & 2 & 0 \\
$\hookrightarrow$ (S) GreedyStepwise & 2 & 3 & 0 \\
$\hookrightarrow$ (S) Ranker & 1 & 0 & 0 \\
JRip & 2 & 2 & 0 \\ 
OneR & 1 & 0 & 0 \\ 
PART & 2 & 2 & 0 \\ 
ZeroR & 0 & 0 & 0 \\ 
DecisionStump & 0 & 0 & 0 \\ 
J48 & 2 & 5 & 0 \\ 
LMT & 1 & 6 & 0 \\ 
REPTree & 2 & 2 & 0 \\ 
RandomForest & 3 & 2 & 0 \\ 
RandomTree & 4 & 4 & 0 \\ \hline

\textbf{Meta-predictors (MP)} & ~ & ~ & ~\\
LWL & 0 & 2 & A \\ 
$\hookrightarrow$ (A) BallTree & 0 & 1 & 2 \\
$\hookrightarrow$ (A) CoverTree & 1 & 1 & 1 \\
$\hookrightarrow$ (A) KDTree & 2 & 1 & 2 \\
$\hookrightarrow$ (A) LinearNNSearch & 0 & 1 & 0 \\
AdaBoostM1 & 2 & 3 & 0 \\ 
AttributeSelectedClassifier & 0 & 0 & S,E \\ 
$\hookrightarrow$ (S) BestFirst & 1 & 1 & 0 \\
$\hookrightarrow$ (S) GreedyStepwise & 2 & 3 & 0 \\
$\hookrightarrow$ (S) Ranker & 1 & 0 & 0 \\
$\hookrightarrow$ (E) CfsSubsetEval & 0 & 2 & 0 \\
$\hookrightarrow$ (E) GainRatioAttributeEval & 0 & 0 & 0 \\
$\hookrightarrow$ (E) InfoGainAttributeEval & 0 & 2 & 0 \\
$\hookrightarrow$ (E) OneRAttributeEval & 2 & 1 & 0 \\
$\hookrightarrow$ (E) WrapperSubsetEval & 0 & 0 & 0 \\
Bagging & 2 & 1 & 0 \\ 
ClassificationViaRegression & 0 & 0 & 0 \\ 
FilteredClassifier & 0 & 0 & 0 \\ 
LogitBoost & 5 & 4 & 0 \\ 
MultiClassClassifier & 1 & 2 & 0 \\ 
RandomCommittee & 1 & 0 & 0 \\ 
RandomSubSpace & 2 & 0 & 0 \\ 
Stacking & 0 & 1 & 0 \\ 
Vote & 0 & 1 & 0 \\ \hline
\end{tabular}
\label{tab:search-space-predictors}
\end{table}

As the datasets we use in our experiments are intended for classification, we have chosen to minimise the classification error averaged over 10 CV folds within the optimisation process (i.e. $k=10$ in Eq.~\ref{eq:MCPS-CASH}).

Two SMBO strategies (SMAC and TPE) have been compared against two baselines (WEKA-Def and random search). The following experimental scenarios were devised:
\begin{itemize}[leftmargin=16px]
\item WEKA-Def: All the predictors and meta-predictors listed in Table~\ref{tab:search-space-predictors} are run using WEKA's default hyperparameter values. Filters are not included in this strategy, although some predictors may perform specific preprocessing steps as part of their default behaviour.
\item Random search: The whole search space is randomly explored allowing 30 CPU core-hours for the process.
\item SMAC and TPE: An initial configuration is randomly selected and then the optimiser is run for 30 CPU core-hours to explore the search space in an intelligent way, allowing for comparison with the random search.
\end{itemize}

In order to compare our results with the ones presented in~\cite{Thornton2013} we have replicated the experimental settings as closely as possible. We have evaluated different optimisation strategies over 21 well-known datasets representing classification tasks (see Table~\ref{tab:datasets}). Each dataset $\mathcal{D} = \{\mathcal{D}_{train}, \mathcal{D}_{test}\}$ has been split into 70\% training and 30\% testing sets, unless partition was already provided. Please note that $\mathcal{D}_{train}$ is then split into 10-folds for Eq.~\ref{eq:MCPS-CASH} and therefore $\mathcal{D}_{test}$ is not used during the optimisation or training process at all (see Figure~\ref{fig:process}).

For each strategy we performed 25 runs with different random seeds within a 30 CPU core-hours optimisation time limit on Intel Xeon E5-2620 six-core 2.00GHz CPU. In the case a configuration step exceeds 30 minutes or 3GB of RAM to evaluate, its evaluation is aborted and not considered further. Once the optimisation process has finished, the returned MCPS is trained using the whole training set $\mathcal{D}_{train}$ and produce predictions for the testing set $\mathcal{D}_{test}$. Please note that this budget limit may imply not finding the optimal solution in the optimisation problem.

Holdout error over $\mathcal{D}_{test}$ is denoted as $\mathcal{E} = \mathcal{L}(Y_{test}, \hat{Y}_{test})$. Random search, SMAC and TPE results have been calculated using the mean of 100,000 bootstrap samples (randomly selecting 4 of the 25 runs and keeping the one with lowest CV error as in original Auto-WEKA paper~\cite{Thornton2013}), while only the lowest errors are reported for WEKA-Def.

\begin{table}
\centering
\caption{Datasets, continuous and categorical attribute count, number of classes, and number of instances.}
\begin{tabular}{l r r r r r}
\hline
\textbf{Dataset} & \textbf{Cont} & \textbf{Disc} & \textbf{Class} & \textbf{Train} & \textbf{Test}\\
\hline
abalone & 7 & 1 & 28 & 2924 & 1253\\
amazon & 10000 & 0 & 50 & 1050 & 450\\
car & 0 & 6 & 4 & 1210 & 518\\
cifar10 & 3072 & 0 & 10 & 50000 & 10000\\
cifar10small & 3072 & 0 & 10 & 10000 & 10000\\
convex & 784 & 0 & 2 & 8000 & 50000\\
dexter & 20000 & 0 & 2 & 420 & 180\\
dorothea & 100000 & 0 & 2 & 805 & 345\\
germancredit & 7 & 13 & 2 & 700 & 300\\
gisette & 5000 & 0 & 2 & 4900 & 2100\\
kddcup09app & 192 & 38 & 2 & 35000 & 15000\\
krvskp & 0 & 36 & 2 & 2238 & 958\\
madelon & 500 & 0 & 2 & 1820 & 780\\
mnist & 784 & 0 & 10 & 12000 & 50000\\
mnistrot & 784 & 0 & 10 & 12000 & 50000\\
secom & 590 & 0 & 2 & 1097 & 470\\
semeion & 256 & 0 & 10 & 1116 & 477\\
shuttle & 9 & 0 & 7 & 43500 & 14500\\
waveform & 40 & 0 & 3 & 3500 & 1500\\
wineqw & 11 & 0 & 11 & 3429 & 1469\\
yeast & 8 & 0 & 10 & 1039 & 445\\
\hline
\end{tabular}
\label{tab:datasets}
\end{table}

\section{Results}
\label{sec:results}

We organised our analysis around the following aspects: a)~usefulness of automatic composition and parametrisation of MCPS; b)~how efficient are Bayesian optimisation approaches (SMAC and TPE) in comparison to random search; c)~impact of significantly extending the search space in the optimisation process; and d)~identification of promising methods for each dataset.

\subsection{Usefulness of automatic composition and parametrisation}

An interesting aspect to analyse is if 30 CPU-core hours of automatic optimisation can beat a quick running of all WEKA classifiers with default hyperparameters. Table~\ref{tab:wekadef-vs-autoweka} shows the results of Auto-WEKA experiments in the NEW search space using RANDOM, SMAC and TPE strategies compared to the WEKA-Def strategy. Not surprisingly, Auto-WEKA has been able to find better results for all datasets ($\delta > 0$). In fact, Auto-WEKA presents a significant improvement in 19 out of 21 datasets.

\begin{table}[t]
\centering
\caption{Holdout error $\mathcal{E}$ (\% missclassification). Lowest errors reported for WEKA-Def and Auto-WEKA, being $\delta$ the difference between the two. Aggregated mean holdout error and standard deviation ($\sigma$) for all Auto-WEKA strategies and search spaces is reported. Each Auto-WEKA run had a 30 CPU core-hours budget. An upward arrow indicates a statistically significant improvement ($p < 0.05$) with respect to best WEKA-Def using one-sample Wilcoxon signed-rank test.}
\setlength\tabcolsep{2pt}
\begin{tabular}{l | r  r  r | r r c }
\hline
 ~ & \multicolumn{1}{c}{\textbf{Best}} & \multicolumn{1}{c}{\textbf{Best}} & \multicolumn{1}{c|}{~} & \multicolumn{1}{c}{\textbf{Mean}} & \multicolumn{1}{c}{~} & ~ \\
 \textbf{Dataset} & \multicolumn{1}{c}{\textbf{WEKA-Def}} & \multicolumn{1}{c}{\textbf{Auto-WEKA}} & \multicolumn{1}{c|}{$\delta$} & \multicolumn{1}{c}{\textbf{Auto-WEKA}} & \multicolumn{1}{c}{$\sigma$} & ~ \\ \hline
abalone & 73.18 & 71.43 & 1.75 & 73.12 & \ci{0.62} & $\uparrow$ \\ 
amazon & 28.44 & 26.67 & 1.77 & 37.07 & \ci{6.61} &  \\ 
car & 0.77 & 0.00 & 0.77 & 0.06 & \ci{0.12} & $\uparrow$ \\ 
cifar10 & 64.27 & 52.28 & 11.99 & 56.21 & \ci{3.47} & $\uparrow$ \\ 
cifar10small & 65.91 & 54.48 & 11.43 & 58.04 & \ci{3.39} & $\uparrow$ \\ 
convex & 25.96 & 18.47 & 7.49 & 22.89 & \ci{2.07} & $\uparrow$ \\ 
dexter & 8.89 & 5.00 & 3.89 & 7.50 & \ci{1.23} & $\uparrow$ \\ 
dorothea & 6.96 & 4.64 & 2.32 & 5.22 & \ci{0.21} & $\uparrow$ \\ 
germancredit & 27.33 & 23.33 & 4.00 & 25.44 & \ci{1.38} & $\uparrow$ \\ 
gisette & 2.81 & 1.95 & 0.86 & 2.33 & \ci{0.24} & $\uparrow$ \\ 
kddcup09app & 1.7405 & 1.6700 & 0.0705 & 1.7339 & \ci{0.0191} & $\uparrow$ \\ 
krvskp & 0.31 & 0.10 & 0.21 & 0.31 & \ci{0.12} &  \\ 
madelon & 21.38 & 15.64 & 5.74 & 17.61 & \ci{1.44} & $\uparrow$ \\ 
mnist & 5.19 & 2.66 & 2.53 & 4.01 & \ci{0.61} & $\uparrow$ \\ 
mnistr & 63.14 & 52.20 & 10.94 & 56.10 & \ci{2.43} & $\uparrow$ \\ 
secom & 8.09 & 7.66 & 0.43 & 7.86 & \ci{0.07} & $\uparrow$ \\ 
semion & 8.18 & 3.98 & 4.20 & 4.93 & \ci{0.48} & $\uparrow$ \\ 
shuttle & 0.0138 & 0.0100 & 0.0038 & 0.0100 & \ci{0.0004} & $\uparrow$ \\ 
waveform & 14.40 & 14.00 & 0.40 & 14.26 & \ci{0.15} & $\uparrow$ \\ 
wineqw & 37.51 & 32.33 & 5.18 & 32.94 & \ci{0.43} & $\uparrow$ \\ 
yeast & 40.45 & 36.40 & 4.05 & 37.73 & \ci{0.74} & $\uparrow$ \\
\hline
\end{tabular}
\label{tab:wekadef-vs-autoweka}
\end{table}

\subsection{Effectiveness of Bayesian optimisation over random search}

Another interesting aspect is to analyse how Bayesian optimisation approaches perform in comparison with just a random search. Table~\ref{tab:random-vs-bayesian} presents the results of Auto-WEKA runs for RANDOM, SMAC and TPE strategies in the NEW search space. SMAC has been able to find significantly better results in all datasets but one. Similarly, TPE outperforms random search in 18 out of 21 datasets.

\begin{table}[htbp]
\centering
\caption{Mean holdout error $\mathcal{E}$ (\% missclassification) and standard deviation ($\sigma$) for RANDOM, SMAC and TPE strategies in the NEW search space. Each Auto-WEKA run had a 30 CPU core-hours budget. An upward arrow indicates a statistically significant improvement ($p < 0.05$) with respect to RANDOM using Wilcoxon signed-rank test.}
\setlength\tabcolsep{3pt}
\begin{tabular}{l | r  r | r  r c | r r c }
\hline
 \textbf{Dataset} & \multicolumn{2}{c|}{\textbf{RANDOM}} & \multicolumn{3}{c|}{\textbf{SMAC}} & \multicolumn{3}{c}{\textbf{TPE}} \\
 ~ & \multicolumn{1}{c}{$\mu$} & \multicolumn{1}{c|}{$\sigma$} & \multicolumn{1}{c}{$\mu$} & \multicolumn{1}{c}{$\sigma$} & ~ & \multicolumn{1}{c}{$\mu$} & \multicolumn{1}{c}{$\sigma$} & ~ \\ \hline
abalone & 72.52 & \ci{0.22} & 72.21 & \ci{0.13} & $\uparrow$ & \textbf{72.01} & \ci{0.21} & $\uparrow$ \\
amazon & 45.75 & \ci{8.69} & \textbf{39.54} & \ci{5.96} & $\uparrow$ & 40.24 & \ci{5.42} & $\uparrow$ \\
car & 0.47 & \ci{0.16} & 0.38 & \ci{0.12} & $\uparrow$ & \textbf{0.21} & \ci{0.10} & $\uparrow$ \\
cifar10 & 58.91 & \ci{3.67} & 56.44 & \ci{2.90} & $\uparrow$ & \textbf{55.60} & \ci{2.77} & $\uparrow$ \\
cifar10small & 60.44 & \ci{4.08} & 57.91 & \ci{2.54} & $\uparrow$ & \textbf{56.57} & \ci{1.79} & $\uparrow$ \\
convex & 25.03 & \ci{1.83} & \textbf{21.88} & \ci{1.37} & $\uparrow$ & 23.19 & \ci{1.12} & $\uparrow$ \\
dexter & 7.54 & \ci{1.25} & 6.42 & \ci{0.42} & $\uparrow$ & \textbf{6.19} & \ci{0.49} & $\uparrow$ \\
dorothea & 6.25 & \ci{0.56} & 5.95 & \ci{0.49} & $\uparrow$ & \textbf{5.92} & \ci{0.53} & $\uparrow$ \\
germancredit & 21.31 & \ci{0.34} & \textbf{19.66} & \ci{0.75} & $\uparrow$ & 19.89 & \ci{0.40} & $\uparrow$ \\
gisette & 2.30 & \ci{0.28} & \textbf{2.21} & \ci{0.31} & $\uparrow$ & 2.35 & \ci{0.23} & ~ \\
kddcup09app & 1.8000 & \ci{0.0000} & \textbf{1.7985} & \ci{0.0036} & $\uparrow$ & 1.8000 & \ci{0.0000} & ~ \\
krvskp & 0.42 & \ci{0.03} & \textbf{0.28} & \ci{0.02} & $\uparrow$ & 0.31 & \ci{0.03} & $\uparrow$ \\
madelon & 19.20 & \ci{1.91} & \textbf{15.61} & \ci{0.70} & $\uparrow$ & 16.03 & \ci{0.41} & $\uparrow$ \\
mnist & 3.78 & \ci{0.57} & \textbf{3.49} & \ci{0.63} & $\uparrow$ & 3.60 & \ci{0.59} & $\uparrow$ \\
mnistr & 58.09 & \ci{1.54} & \textbf{55.75} & \ci{2.44} & $\uparrow$ & 57.17 & \ci{2.13} & $\uparrow$ \\
secom & 5.85 & \ci{0.35} & 6.00 & \ci{0.09} & ~ & 5.85 & \ci{0.41} & ~ \\
semeion & 4.82 & \ci{0.35} & 4.48 & \ci{0.47} & $\uparrow$ & \textbf{4.28} & \ci{0.37} & $\uparrow$ \\
shuttle & 0.0109 & \ci{0.0028} & \textbf{0.0103} & \ci{0.0016} & $\uparrow$ & 0.0107 & \ci{0.0026} & $\uparrow$ \\
waveform & 12.50 & \ci{0.09} & \textbf{12.33} & \ci{0.08} & $\uparrow$ & 12.43 & \ci{0.04} & $\uparrow$ \\
wineqw & 33.08 & \ci{0.24} & \textbf{32.64} & \ci{0.27} & $\uparrow$ & 32.67 & \ci{0.18} & $\uparrow$ \\
yeast & 37.16 & \ci{0.34} & 36.50 & \ci{0.35} & $\uparrow$ & \textbf{36.17} & \ci{0.23} & $\uparrow$ \\
\hline
\end{tabular}
\label{tab:random-vs-bayesian}
\end{table}

\subsection{Impact of extending the search space}

As shown in Table~\ref{tab:search-spaces}, the size between search spaces vary a lot: PREV with over 21 thousands; NEW with over 2 million; and FULL with over 812 billion possible solutions. We are interested on analysing the impact of extending the search space in the classification performance for each strategy.

Table~\ref{tab:test-error} presents the results of RANDOM, SMAC and TPE over the three different search spaces. In the majority of cases (52 of 63), the MCPSs found in the NEW search space achieve significantly better results than in the smaller search space PREV. In 32 out of 63 cases, the FULL search space also gets significantly better performance than in PREV. However, finding good MCPS within the same time budget (30 CPU-core hours) is more challenging due to a large increase in the search space size~\cite{Hoos2014}. As an example, consider Figure~\ref{fig:madelon-trajectories} where the evolution of the best solution for `madelon' dataset and SMAC strategy is represented over time for each of the 25 runs. Comparing Figures~\ref{fig:madelon-trajectories}-a) and b) we can see that the rate of convergence is much higher in the smaller space (denoted as NEW). Nevertheless, the overall best-performing model for `madelon' was found in the FULL space as seen in Table~\ref{tab:MCPS-configurations}.

The way in which the search space is extended can have a considerable impact on the accuracy of the MCPS found. Additional hyperparameters allowing for extra tuning flexibility (PREV to NEW) improved the performance in most of the cases. However, adding more transitions to the MCPS (NEW to FULL) does not seem to help on average, given the same CPU time limit. Nevertheless, the best MCPSs found in the FULL search space for 13 out of 28 datasets have better or comparable performance to the best solutions found in the NEW space as shown in Table~\ref{tab:MCPS-configurations}.

\begin{table*}
\centering
\caption{Mean holdout error $\mathcal{E}$ (\% missclassification) and standard deviation ($\sigma$) for RANDOM, SMAC and TPE with a 30 CPU core-hours budget per run. PREV columns contain the values reported in~\cite{Thornton2013}, while NEW and FULL columns contain the results for the search spaces described in Section~\ref{sec:methodology}. Boldfaced values indicate the lowest mean classification error for each dataset. An upward arrow indicates a statistically significant improvement ($p < 0.05$) with respect to PREV search space using one-sample Wilcoxon signed-rank test.}
\setlength\tabcolsep{2pt}
\begin{tabular}{l | r r r r r r r | r r r r r r r | r r r r r r r}
\hline
\multicolumn{1}{l}{\textbf{dataset}} & \multicolumn{7}{c}{\textbf{RANDOM}} & \multicolumn{7}{c}{\textbf{SMAC}} & \multicolumn{7}{c}{\textbf{TPE}} \\ \cline{2-22}

~ & \multicolumn{1}{c}{PREV} & \multicolumn{3}{c}{NEW} & \multicolumn{3}{c}{FULL} & \multicolumn{1}{c}{PREV} & \multicolumn{3}{c}{NEW} & \multicolumn{3}{c}{FULL} & \multicolumn{1}{c}{PREV} & \multicolumn{3}{c}{NEW} & \multicolumn{3}{c}{FULL} \\ \hline

~ & \multicolumn{1}{c}{$\mu$} & \multicolumn{1}{c}{$\mu$} & \multicolumn{1}{c}{$\sigma$} & ~ & \multicolumn{1}{c}{$\mu$} & \multicolumn{1}{c}{$\sigma$} & ~ & \multicolumn{1}{c}{$\mu$} & \multicolumn{1}{c}{$\mu$} & \multicolumn{1}{c}{$\sigma$} & ~ & \multicolumn{1}{c}{$\mu$} & \multicolumn{1}{c}{$\sigma$} & ~ & \multicolumn{1}{c}{$\mu$} & \multicolumn{1}{c}{$\mu$} & \multicolumn{1}{c}{$\sigma$} & ~ & \multicolumn{1}{c}{$\mu$} & \multicolumn{1}{c}{$\sigma$} & ~ \\ \hline

abalone & 74.88 & \textbf{72.92} & \ci{0.78} & $\uparrow$ & 73.76 & \ci{0.48} & $\uparrow$ & 73.51 & 73.40 & \ci{0.50} & $\uparrow$ & 73.16 & \ci{0.66} & $\uparrow$ & 72.94 & 73.03 & \ci{0.40} & ~ & 73.26 & \ci{0.54} & ~ \\

amazon & 41.11 & 39.18 & \ci{9.17} & $\uparrow$ & 56.65 & \ci{15.60} & ~ & \textbf{33.99} & 36.28 & \ci{4.41} & ~ & 49.11 & \ci{16.00} & ~ & 36.59 & 35.71 & \ci{4.60} & $\uparrow$ & 54.06 & \ci{11.79} & ~ \\

car & 0.01 & 0.13 & \ci{0.15} & ~ & 1.84 & \ci{1.04} & ~ & 0.40 & 0.05 & \ci{0.11} & $\uparrow$ & 0.20 & \ci{0.23} & $\uparrow$ & 0.18 & 0.01 & \ci{0.04} & $\uparrow$ & 0.05 & \ci{0.12} & $\uparrow$ \\

cifar10 & 69.72 & 58.21 & \ci{3.65} & $\uparrow$ & 66.59 & \ci{3.35} & $\uparrow$ & 61.15 & 55.52 & \ci{2.90} & $\uparrow$ & 68.55 & \ci{3.71} & ~ & 66.01 & \textbf{54.88} & \ci{2.82} & $\uparrow$ & 65.30 & \ci{4.29} & $\uparrow$ \\

cifar10small & 66.12 & 59.85 & \ci{4.32} & $\uparrow$ & 71.61 & \ci{4.55} & ~ & 56.84 & 57.85 & \ci{2.62} & ~ & 71.82 & \ci{8.07} & ~ & 57.01 & \textbf{56.43} & \ci{1.74} & $\uparrow$ & 68.40 & \ci{7.00} & ~ \\

convex & 31.20 & 24.76 & \ci{1.90} & $\uparrow$ & 33.02 & \ci{5.32} & ~ & 23.17 & \textbf{21.31} & \ci{1.48} & $\uparrow$ & 24.52 & \ci{2.17} & ~ & 25.59 & 22.62 & \ci{1.03} & $\uparrow$ & 30.57 & \ci{3.74} & ~ \\

dexter & 9.18 & 8.27 & \ci{1.19} & $\uparrow$ & 11.27 & \ci{2.70} & ~ & 7.49 & 7.31 & \ci{0.85} & $\uparrow$ & 8.13 & \ci{4.02} & ~ & 8.89 & \textbf{6.90} & \ci{1.18} & $\uparrow$ & 8.00 & \ci{1.82} & $\uparrow$ \\

dorothea & 5.22 & 5.27 & \ci{0.15} & ~ & 5.37 & \ci{0.40} & ~ & 6.21 & 5.12 & \ci{0.27} & $\uparrow$ & 5.49 & \ci{0.35} & $\uparrow$ & 6.15 & 5.25 & \ci{0.16} & $\uparrow$ & 5.12 & \ci{0.31} & $\uparrow$ \\

germancredit & 29.03 & \textbf{25.40} & \ci{1.56} & $\uparrow$ & 26.87 & \ci{1.42} & $\uparrow$ & 28.24 & 25.43 & \ci{1.55} & $\uparrow$ & 26.67 & \ci{1.20} & $\uparrow$ & 27.54 & 25.49 & \ci{0.95} & $\uparrow$ & 26.63 & \ci{1.24} & $\uparrow$ \\

gisette & 4.62 & 2.28 & \ci{0.24} & $\uparrow$ & 3.36 & \ci{1.39} & $\uparrow$ & \textbf{2.24} & 2.35 & \ci{0.26} & ~ & 2.74 & \ci{1.57} & ~ & 3.94 & 2.37 & \ci{0.19} & $\uparrow$ & 2.86 & \ci{0.46} & $\uparrow$ \\

kddcup09app & 1.7400 & \textbf{1.7214} & \ci{0.0296} & $\uparrow$ & 1.7403 & \ci{0.1103} & ~ & 1.7358 & 1.7400 & \ci{0.0000} & ~ & 1.7400 & \ci{0.0000} & ~ & 1.7381 & 1.7400 & \ci{0.0000} & ~ & 1.7400 & \ci{0.0000} & ~ \\

krvskp & 0.58 & 0.34 & \ci{0.10} & $\uparrow$ & 0.39 & \ci{0.15} & $\uparrow$ & 0.31 & \textbf{0.23} & \ci{0.11} & $\uparrow$ & 0.39 & \ci{0.07} & ~ & 0.54 & 0.36 & \ci{0.09} & $\uparrow$ & 0.33 & \ci{0.08} & $\uparrow$ \\

madelon & 24.29 & 19.11 & \ci{1.30} & $\uparrow$ & 23.80 & \ci{4.53} & $\uparrow$ & 21.56 & \textbf{16.80} & \ci{0.95} & $\uparrow$ & 17.16 & \ci{3.68} & $\uparrow$ & 21.12 & 16.91 & \ci{0.52} & $\uparrow$ & 17.59 & \ci{2.59} & $\uparrow$ \\

mnist & 5.05 & 4.00 & \ci{0.68} & $\uparrow$ & 10.93 & \ci{5.57} & ~ & \textbf{3.64} & 4.08 & \ci{0.38} & ~ & 10.57 & \ci{6.83} & ~ & 12.28 & 3.96 & \ci{0.73} & $\uparrow$ & 12.32 & \ci{9.13} & ~ \\

mnistr & 66.40 & 57.15 & \ci{1.80} & $\uparrow$ & 65.89 & \ci{4.83} & $\uparrow$ & 57.04 & \textbf{54.84} & \ci{2.51} & $\uparrow$ & 65.48 & \ci{5.62} & ~ & 70.20 & 56.30 & \ci{2.33} & $\uparrow$ & 63.90 & \ci{5.36} & $\uparrow$ \\

secom & 8.03 & 7.88 & \ci{0.05} & $\uparrow$ & 7.87 & \ci{0.00} & $\uparrow$ & 8.01 & 7.87 & \ci{0.01} & $\uparrow$ & 7.87 & \ci{0.00} & $\uparrow$ & 8.10 & \textbf{7.84} & \ci{0.08} & $\uparrow$ & 7.87 & \ci{0.00} & $\uparrow$ \\

semeion & 6.10 & \textbf{4.78} & \ci{0.59} & $\uparrow$ & 8.20 & \ci{1.40} & ~ & 5.08 & 5.09 & \ci{0.49} & ~ & 5.46 & \ci{0.77} & ~ & 8.26 & 4.91 & \ci{0.21} & $\uparrow$ & 6.31 & \ci{1.12} & $\uparrow$ \\

shuttle & 0.0157 & 0.0100 & \ci{0.0005} & $\uparrow$ & 0.0217 & \ci{0.0225} & ~ & 0.0130 & 0.0100 & \ci{0.0005} & $\uparrow$ & 0.0075 & \ci{0.0054} & $\uparrow$ & 0.0145 & 0.0100 & \ci{0.0003} & $\uparrow$ & \textbf{0.0077} & \ci{0.0029} & $\uparrow$ \\

waveform & 14.27 & 14.26 & \ci{0.16} & $\uparrow$ & 14.28 & \ci{0.46} & ~ & 14.42 & 14.17 & \ci{0.09} & $\uparrow$ & \textbf{13.99} & \ci{0.30} & $\uparrow$ & 14.23 & 14.34 & \ci{0.14} & ~ & 14.05 & \ci{0.25} & $\uparrow$ \\

wineqw & 34.41 & 32.99 & \ci{0.58} & $\uparrow$ & 36.64 & \ci{1.48} & ~ & 33.95 & 32.89 & \ci{0.31} & $\uparrow$ & 34.14 & \ci{0.76} & ~ & 33.56 & \textbf{32.93} & \ci{0.33} & $\uparrow$ & 34.09 & \ci{0.44} & ~ \\

yeast & 43.15 & 37.68 & \ci{0.75} & $\uparrow$ & 40.86 & \ci{1.05} & $\uparrow$ & 40.67 & \textbf{37.60} & \ci{0.64} & $\uparrow$ & 39.01 & \ci{0.78} & $\uparrow$ & 40.10 & 37.89 & \ci{0.78} & $\uparrow$ & 38.91 & \ci{0.66} & $\uparrow$ \\

\hline
\end{tabular}
\label{tab:test-error}
\end{table*}

\begin{figure*}
\centering
\includegraphics[width=0.45\linewidth]{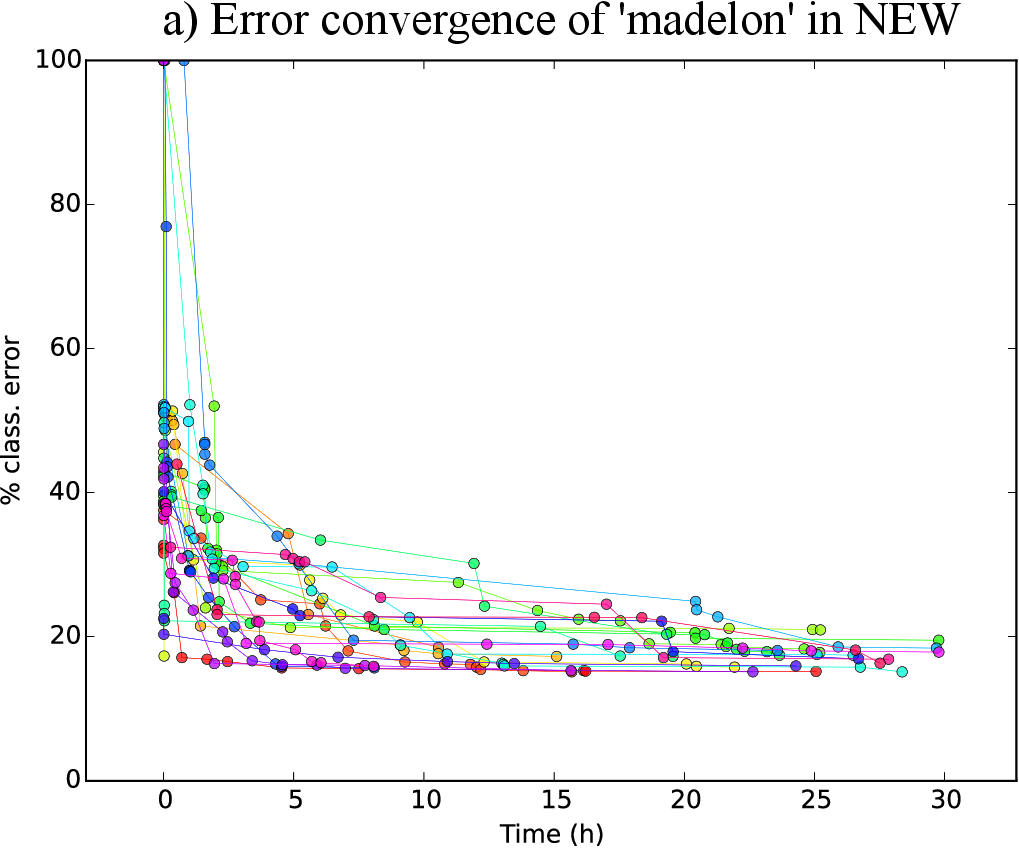}
\hspace{\fill}
\includegraphics[width=0.45\linewidth]{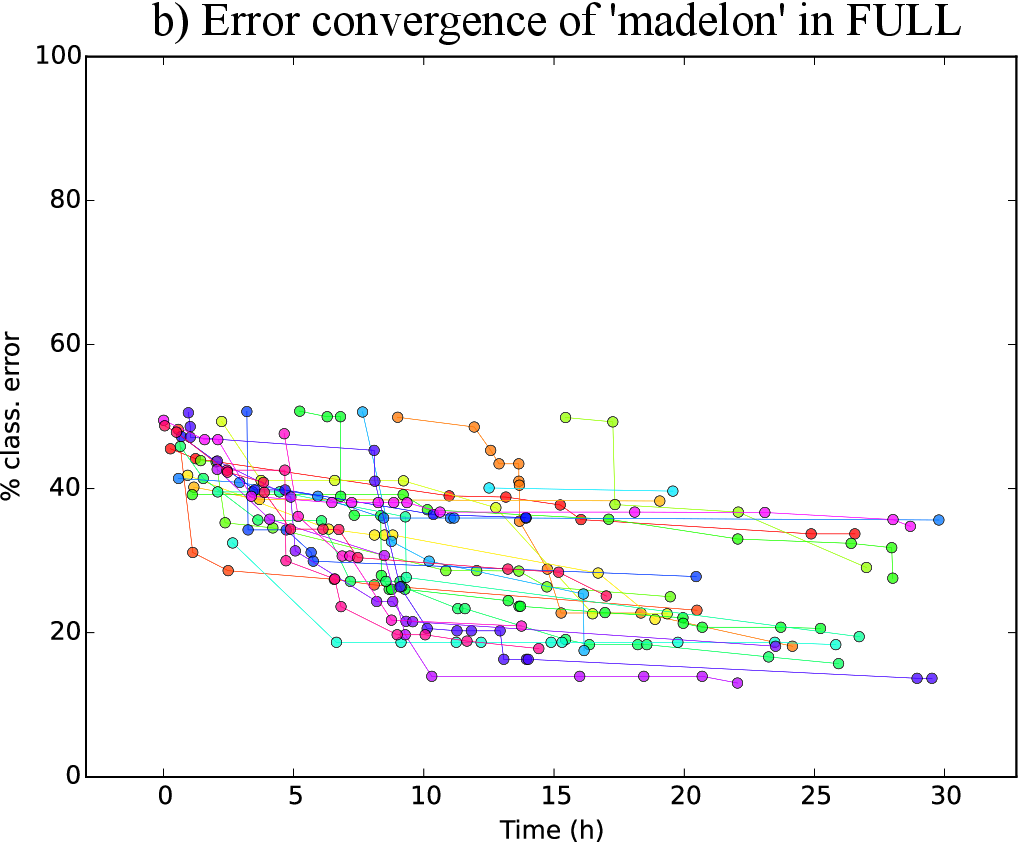}
\caption{10-fold CV error of best solutions found over time for `madelon' dataset and SMAC strategy in a) NEW and b) FULL search spaces.}
\label{fig:madelon-trajectories}
\end{figure*}

\subsection{Identifying promising configurations}

The best MCPSs found for each dataset are reported in Table~\ref{tab:MCPS-configurations}, where each row represents a sequence of data transformations and predictive models as explained in Section~\ref{sec:methodology}.
The solutions found for different datasets are quite diverse, and they often also vary a lot across the 25 random runs performed for each dataset. In order to better understand the observed differences in the MCPSs found we have also measured the average pairwise similarity of the 25 MCPSs found for each dataset and the variance of their performances (see Figure~\ref{fig:similarity-vs-variance}). To calculate the similarity between configurations a weighted sum of Hamming distances given by
\begin{equation}
  d(\theta_a,\theta_b) = 1 - \frac{\sum_{i=1}^{n}{(\omega_i \cdot \delta_i)}}{\sum_{i=1}^{n}{\omega_i}}
\end{equation}
is used, where $\theta_a$ and $\theta_b$ are MCPSs with $N$ transitions, $\omega_i \in \Omega$ is the weight for the $i$th transition and $\delta_i$ is the Hamming distance (a standard measure of string dissimilarity) of components at position $i$.

Weights have been fixed manually to $\Omega = \{2, 1.5\}$ in the NEW search space and $\Omega = \{1, 1, 1, 1, 1, 2, 1.5\}$ in the FULL search space. One could however set the weights in a different way depending on what components are believed to be more relevant. In this case, preprocessing transitions have the same weight while both predictors and meta-predictors have higher weights because of their importance~\cite{Hoos2014}.

\begin{table*}
\centering
\caption{Best MCPS for each dataset in NEW and FULL spaces and its holdout error. MV = missing value replacement, OU = outlier detection and removal, TR = transformation, DR = dimensionality reduction, SA = sampling. }
\begin{tabular}{l l l l l l l l l r}
\hline
\textbf{dataset} & \textbf{space} & \textbf{MV} & \textbf{OU} & \textbf{TR} & \textbf{DR} & \textbf{SA} & \textbf{predictor} & \textbf{meta-predictor} & \multicolumn{1}{c}{$\mathcal{E}$}\\
\hline
\multirow{2}{*}{abalone} & NEW & - & - & - & - & - & MLP & RandomCommittee & 71.43 \\
                         & FULL & Median & - & Center & RandomSubset & Resample & Logistic & Bagging & 72.39 \\ \hline
        
\multirow{2}{*}{amazon} & NEW & - & - & - & - & - & SimpleLogistic & RandomSubSpace & 26.67 \\
                        & FULL & Min & - & Normalize & RandomSubset & - & NaiveBayesMult. & RandomSubSpace & 20.89 \\ \hline

\multirow{2}{*}{car} & NEW & - & - & - & - & - & SMO & MultiClassClassifier & 0.00 \\
                     & FULL & - & - & Standardize & - & Resample & SMO & AdaBoostM1 & 0.00 \\ \hline

\multirow{2}{*}{cifar10} & NEW & - & - & - & - & - & RandomForest & MultiClassClassifier & 52.28 \\
                         & FULL & - & - & - & - & Resample & RandomTree & Bagging & 59.63 \\ \hline

\multirow{2}{*}{cifar10small} & NEW & - & - & - & - & - & RandomTree & MultiClassClassifier & 54.48 \\
                              & FULL & - & - & - & RandomSubset & - & RandomTree & AdaBoostM1 & 59.97 \\ \hline

\multirow{2}{*}{convex} & NEW & - & - & - & - & - & RandomForest & AdaBoostM1 & 18.47 \\
                        & FULL & - & - & Center & - & Resample & RandomTree & AdaBoostM1 & 22.97 \\ \hline

\multirow{2}{*}{dexter} & NEW & - & - & - & - & - & DecisionStump & AdaBoostM1 & 5.00 \\
                        & FULL & - & - & - & - & Resample & VotedPerceptron & AdaBoostM1 & 5.00 \\ \hline

\multirow{2}{*}{dorothea} & NEW & - & - & - & - & - & OneR & RandomSubSpace & 4.64 \\
                          & FULL & - & - & Standardize & - & - & REPTree & LogitBoost & 4.64 \\ \hline

\multirow{2}{*}{germancredit} & NEW & - & - & - & - & - & LMT & Bagging & 23.33 \\
                              & FULL & Zero & - & Standardize & RandomSubset & - & LMT & Bagging & 24.33 \\ \hline

\multirow{2}{*}{gisette} & NEW & - & - & - & - & - & NaiveBayes & LWL & 1.95 \\
                         & FULL & - & - & - & - & - & VotedPerceptron & RandomSubSpace & 1.52 \\ \hline

\multirow{2}{*}{kddcup09app} & NEW & - & - & - & - & - & ZeroR & LWL & 1.67 \\
                             & FULL & - & IQR & Standardize & Attr. Selection & Reservoir & BayesNet & MultiClassClassifier & 1.74 \\ \hline

\multirow{2}{*}{krvskp} & NEW & - & - & - & - & - & JRip & AdaBoostM1 & 0.10 \\
                        & FULL & - & - & Normalize & - & - & JRip & AdaBoostM1 & 0.21 \\ \hline

\multirow{2}{*}{madelon} & NEW & - & - & - & - & - & REPTree & RandomSubSpace & 15.64 \\
                         & FULL & - & - & - & PCA & - & IBk & LogitBoost & 12.82 \\ \hline

\multirow{2}{*}{mnist} & NEW & - & - & - & - & - & SMO & MultiClassClassifier & 2.66 \\
                       & FULL & Zero & - & Center & - & - & J48 & AdaBoostM1 & 5.15 \\ \hline

\multirow{2}{*}{mnistr} & NEW & - & - & - & - & - & RandomForest & RandomCommittee & 52.20 \\
                        & FULL & Zero & - & Normalize & - & - & BayesNet & RandomSubSpace & 56.33 \\ \hline

\multirow{2}{*}{secom} & NEW & - & - & - & - & - & J48 & AdaBoostM1 & 7.66 \\
                       & FULL & - & - & Standardize & - & Reservoir & ZeroR & FilteredClassifier & 7.87 \\ \hline

\multirow{2}{*}{semeion} & NEW & -  & - & - & - & - & NaiveBayes & LWL & 3.98 \\
                         & FULL & EM & - & - & PCA & - & SMO & FilteredClassifier & 4.61 \\ \hline

\multirow{2}{*}{shuttle} & NEW & - & - & - & - & - & RandomForest & AdaBoostM1 & 0.01\\
                         & FULL & - & - & Center & - & Resample & REPTree & AdaBoostM1 & 0.01 \\ \hline

\multirow{2}{*}{waveform} & NEW & - & - & - & - & - & SMO & RandomSubSpace & 14.00 \\
                          & FULL & - & IQR & Normalize & - & - & SMO & Attr.SelectedClassifier & 13.40 \\ \hline

\multirow{2}{*}{wineqw} & NEW & - & - & - & - & - & RandomForest & AdaBoostM1 & 32.33 \\
                        & FULL & Mean & - & Wavelet & Attr.Selection & - & IBk & RandomSubSpace & 33.42 \\ \hline

\multirow{2}{*}{yeast} & NEW & - & - & - & - & - & RandomForest & Bagging & 36.40 \\
                       & FULL & - & - & Normalize & - & - & RandomTree & Bagging & 38.20 \\ \hline

\end{tabular}
\label{tab:MCPS-configurations}
\end{table*}

It is worth mentioning that most of the MCPSs found for the `waveform' dataset include a missing value replacement method even though there are no missing values in this dataset and therefore it doesn't have any effect on the data or the classification performance. The presence of such an unnecessary component likely stems from the fact that selecting a method for replacing missing values at random has a prior probability of 0.75 (i.e. 3 out of 4 possible actions as seen in Table~\ref{tab:search-space-predictors}) which means that it can be selected when randomly initialising the configurations of MCPSs to start from and using the search method which does not penalise unnecessary elements in the data processing chains. However, it is not the case with other components like `Transformation' in which although the prior probability of selecting one of the available transformation methods is 5/6, selecting an appropriate method has a potential impact on the performance and therefore better transformation methods tend to be retained in the found solutions.

\begin{figure}
\centering
\includegraphics[width=\linewidth]{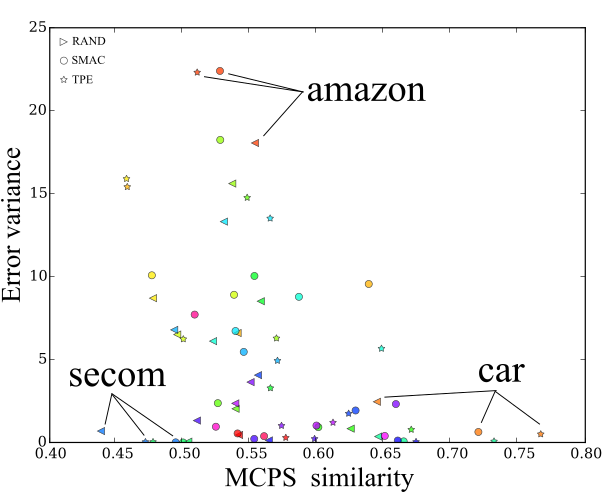}
\caption{Error variance vs. MCPS similarity in FULL search space}
\label{fig:similarity-vs-variance}
\end{figure}

For illustrative purposes we have selected three interesting cases from Figure~\ref{fig:similarity-vs-variance} for a more detailed analysis:
\begin{itemize}[leftmargin=16px]
\item \textbf{Low error variance and high MCPS similarity.} Most of the best solutions found follow a very similar sequence of methods. Therefore similar classification performance is to be expected. For example, a repeated sequence in `car' dataset with TPE optimisation is MultiLayerPerceptron (13/25) $\rightarrow$ AdaBoostM1 (22/25).
\item \textbf{Low error variance and low MCPS similarity.} Despite having different solutions, classification performance in a group of analysed datasets does not vary much. This can mean that the classification problem is not difficult and a range of different MCPSs can perform quite well on it. This is for instance the case of the solutions found for the `secom' and `kddcup09app' datasets.
\item \textbf{High error variance and low MCPS similarity.} In such cases, there are many differences between both the best MCPSs found and their classification performances. For instance, it is the case of `amazon' dataset for which a high error variance was observed in all of the optimisation strategies (see Figure~\ref{fig:similarity-vs-variance}). We believe such difference likely results from a combination of difficulty of the classification task (i.e.~high input dimensionality, large number of classes and a relatively small number of training samples) and/or insufficient exploration from the random starting configuration in a very large space.
\end{itemize}

\section{Application to process industry}
\label{sec:process-industry}

\begin{figure}[!htbp]
\centering
\includegraphics[width=\linewidth]{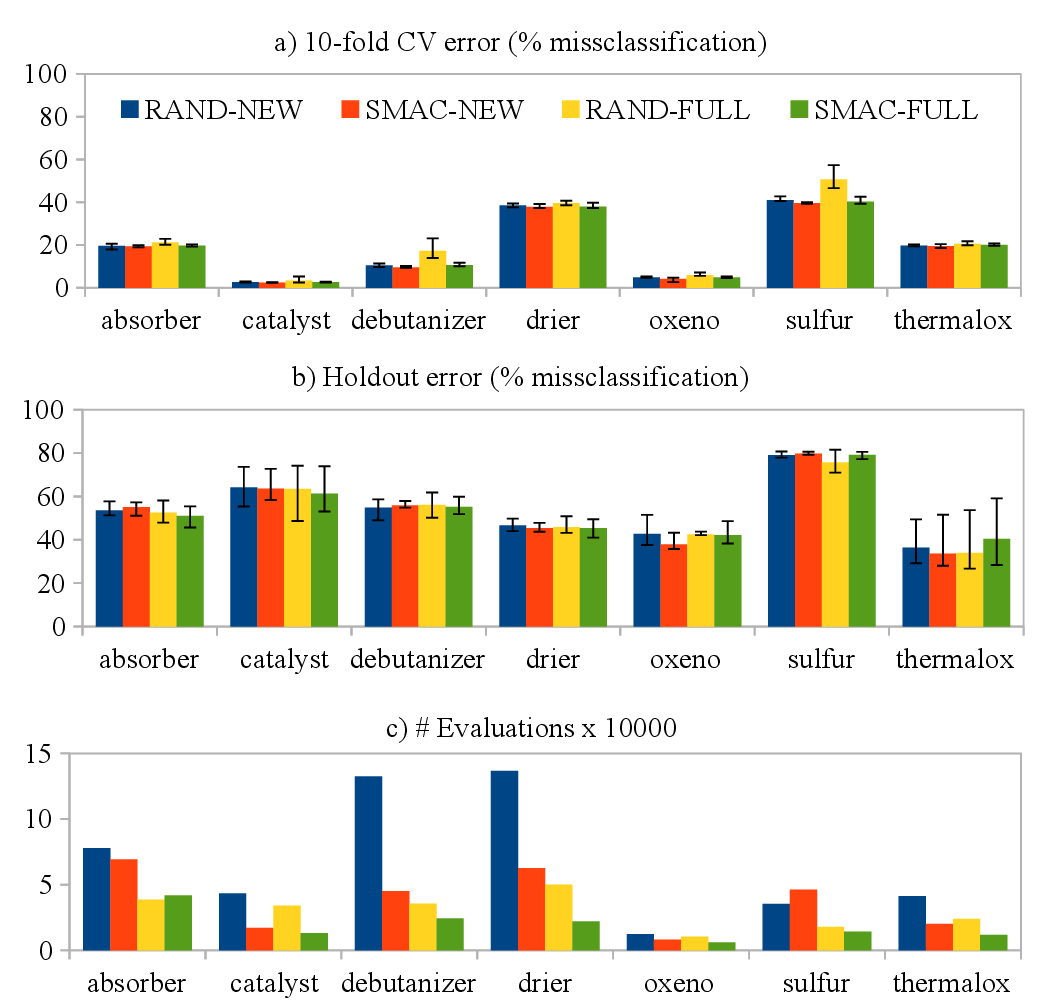}
\caption{a) Mean 10-fold CV error $\epsilon$ and b) holdout error $\mathcal{E}$ with 95\% bootstrap confidence intervals for process industry datasets. c) Total number of evaluations per dataset and strategy.}
\label{fig:process-industry-results}
\end{figure}

One motivation for automating the composition and optimisation of MCPSs was the need for speeding up the process of developing soft-sensors~\cite{Kadlec2009a}, which are predictive models based on easy-to-measure quantities used in the process industry. Main applications of soft-sensors are online prediction~\cite{Zhou2014}, process monitoring~\cite{Zhou2016} and fault detection~\cite{GeMing2008}. The most popular methods for process monitoring include Principal Component Analysis (PCA \cite{Jolliffe2002}) in a combination with a predictor, Multi-Layer Perceptron (MLP \cite{Qin1997}), Radial Basis Function (RBF \cite{Wang2006}) and Self Organizing Map (SOM \cite{Alhoniemi1999}). \cite{Kadlec2009a} show that there are indeed dozens of methods to build soft sensors and each of them with various hyperparameters. Our experience in this field comes from past involvement in multiple projects with chemical engineering companies~\cite{Kadlec2011,MartinSalvador2014,MartinSalvador2016a}. 

Raw data from chemical plants usually requires a considerable preprocessing and modelling effort~\cite{Lin2007,Budka2014}. Although some WEKA predictors include inner preprocessing such as removal of missing values or normalisation, as shown previously in~\cite{MartinSalvador2016}, many datasets need additional preprocessing to build effective predictive models. The fixed order of preprocessing nodes in the FULL search space has not been set arbitrarily -- it follows the preprocessing guidelines that are common in process industry when developing predictive models (see e.g.~\cite{Budka2014, Kadlec2009a}).

\begin{table*}[!ht]
\centering
\caption{Best MCPS for process industry datasets in NEW and FULL spaces, holdout error $\mathcal{E}$ and difference with baseline $\delta$ ($\uparrow$ indicates an improvement). MV = missing value replacement, OU = outlier removal, TR = transformation, DR = dim. reduction, SA = sampling.}
\begin{tabular}{l l l l l l l l l r r}
\hline
\textbf{dataset} & \textbf{space} & \textbf{MV} & \textbf{OU} & \textbf{TR} & \textbf{DR} & \textbf{SA} & \textbf{predictor} & \textbf{meta-predictor} & \multicolumn{1}{c}{$\mathcal{E}$} & \multicolumn{1}{c}{$\delta$}\\
\hline
\multirow{2}{*}{absorber} & NEW & - & - & - & - & - & REPTree & AdaBoostM1 & 51.04 & -14.79 \\
                          & FULL & - & - & Wavelet & RandomSubset & - & KStar & Bagging & 45.63 & -9.38 \\ \hline
                       
\multirow{2}{*}{catalyst} & NEW & - & - & - & - & - & LMT & AdaBoostM1 & 55.34 & -22.90 \\
                          & FULL & - & - & Wavelet & RandomSubset & - & JRip & FilteredClassifier & 48.64 & -16.20 \\ \hline

\multirow{2}{*}{debutanizer} & NEW & - & - & - & - & - & LMT & AdaBoostM1 & 49.03 & $\uparrow$ 3.20\\
                             & FULL & Median & - & Wavelet & - & - & REPTree & FilteredClassifier & 50.14 & $\uparrow$ 2.09\\ \hline
                       
\multirow{2}{*}{drier} & NEW & - & - & - & - & - & JRip & Bagging & 43.72 & $\uparrow$ 7.37\\
                       & FULL & - & IQR & Normalize & - & - & Logistic & Bagging & 40.98 & $\uparrow$ 10.11 \\ \hline

\multirow{2}{*}{oxeno} & NEW & - & - & - & - & - & JRip & AdaBoostM1 & 35.74 & -2.22 \\
                       & FULL & - & - & Standardize & - & - & JRip & RandomSubSpace & 38.24 & -4.74\\ \hline
                             
\multirow{2}{*}{sulfur} & NEW & - & - & - & - & - & PART & Bagging & 77.91 & $\uparrow$ 1.75 \\
                        & FULL & - & - & Wavelet & RandomSubset & - & JRip & FilteredClassifier & 70.96 & $\uparrow$ 8.70 \\ \hline
                       
\multirow{2}{*}{thermalox} & NEW & - & - & - & - & - & Logistic & MultiClassClassifier & 28.01 & $\uparrow$ 19.51 \\
                           & FULL & - & - & Wavelet & - & - & MLP & FilteredClassifier & 26.71 & $\uparrow$ 20.81 \\ \hline

\end{tabular}
\label{tab:MCPS-configurations-industry}
\end{table*}

We have carried out an experimental analysis on 7 datasets representing process monitoring tasks of real chemical processes (i.e. classification of 3 process states -- low, normal and high). Four of these datasets have been made available by Evonik Industries as part of the collaboration within the INFER project \cite{Musial2013}, and have been extensively used in previous studies \cite{Kadlec2009b, Budka2014, Bakirov2015}:
\begin{itemize}
    \item `absorber' which contains 38 continuous attributes from an absorption process. No additional information has been provided apart from this being a regression task;
    \item `drier' with 19 continuous features from physical sensors (i.e. temperature, pressure and humidity) and the target value is the residual humidity of the product \cite{Kadlec2009b};
    \item `oxeno' which contains 71 continuous attributes also from physical sensors and a target variable which is the product concentration measured in the laboratory \cite{Budka2014}; and
    \item `thermalox' which has 38 attributes from physical sensors and the two target values are concentrations of $NO_x$ and $SO_x$ in the exhaust gases \cite{Kadlec2009b}.
\end{itemize}
Due to confidentiality reasons the datasets listed above cannot be published. However, 3 additional publicly available datasets from the same domain have also been used in the experiments: 
\begin{itemize}
    \item `catalyst' with 14 attributes, where the task is to predict the activity of a catalyst in a multi-tube reactor \cite{Kadlec2011};
    \item `debutanizer' which has 7 attributes (temperature, pressure and flow measurements of a debutanizer column) and where the target value is the concentration of butane at the output of the column \cite{Fortuna2005}; and
    \item the `sulfur' recovery unit, which is a system for removing environmental pollutants from acid gas streams before they are released into the atmosphere \cite{Fortuna2003}. The washed out gases are transformed into sulfur. The dataset has five input features (flow measurements) and two target values: concentration of $H_{2}S$ and $SO_2$.
\end{itemize}

The results presented in Figure~\ref{fig:process-industry-results} show that including such preprocessing steps has allowed, on average, to find better MCPSs than in the NEW search space for selected datasets. Although the difference is small, that implies that considerably extending the search space does not only have major negative effects but also can be positive for the predictive performance on these datasets. It is interesting to highlight that random search was able to find the MCPSs with lowest holdout error in half of the cases, but on the other hand also presents a higher error variance. We believe that the reason behind this is that random search has evaluated more models than SMAC for almost all the runs (see Figure~\ref{fig:process-industry-results}-c)). This suggests that random search -- which evaluates more potential solutions and explores more regions of the search space -- can, and in these few cases has found better solutions than SMAC.

The best MCPSs found for the chemical datasets are shown at the end of Table~\ref{tab:MCPS-configurations-industry}. These solutions outperform the four most popular methods for building soft sensors for process monitoring (PCA, MLP, RBF and SOM) in 4 out of 7 datasets (see $\delta$ in Table~\ref{tab:MCPS-configurations-industry}). Also, MCPSs including an attribute selection step have a considerable improvement of performance with respect to NEW (e.g. `absorber', `catalyst' and `sulfur').

We also can see in Figure~\ref{fig:process-industry-results} that there is a large difference between the CV error and the holdout test error in some of these datasets (e.g. $\epsilon = 2.60\%$ to $\mathcal{E} = 61.27\%$ in `catalyst'). This is due to the evolving nature of some chemical processes over time. The test set (last 30\% of samples) can be significantly different from the initial 70\% of data used for training the MCPS. We have shown in \cite{MartinSalvador2016a} how it is possible to adapt these MCPSs, using the proposed automatic composition and optimisation approach repeatedly, without any human intervention, when there are changes in data.


\section{Conclusions and future work}
\label{sec:conclusion}

In this work automatic composition and optimisation of MCPSs has been addressed as a search problem. In particular, we proposed extending the CASH problem to MCPS represented as Petri nets, where transitions are data processing methods with hyperparameters. To apply this approach to real problems, we have extended Auto-WEKA to support any type of preprocessing methods available in WEKA and thus form MCPSs. In an extensive experimental analysis using 21 publicly available datasets and 7 challenging datasets from the process industry, we have demonstrated that it is indeed possible to find feasible MCPSs to solve predictive problems. Results have indicated that Sequential Model-Based Optimisation (SMBO) strategies perform better than random search given the same time for optimisation in the majority of the analysed datasets. As a consequence, we can considerably reduce the human effort and speed up the data mining process.

Based on the error variance among different runs of the extended Auto-WEKA on the same dataset and the similarity of the MCPSs found, we have identified three interesting cases which may happen when finding the best solution using multiple starting points in the search space: (1) convergence towards very similar solutions; (2) small error variation between different solutions; and (3) very different solutions. From the practical point of view, a single composition and optimisation run should be sufficient when dealing with datasets falling into the first two categories. On the other hand, it is clear that multiple composition and optimisation runs with different starting points are needed for datasets in case (3). The challenge is to identify a priori to which category any of the considered datasets belong. This is a promising future work direction, with a potential of further time and computational cost savings. A similar challenge is to define a priori or in a flexible, dynamic manner a time budget for any particular dataset.

In contrast to the collection of datasets presented in~\cite{Thornton2013} (and also evaluated in this paper), data distribution of the evaluated process industry datasets is changing over time. In these cases there is a need to adapt the optimised MCPSs following the changing environment. The first approach to deal with concept drift while optimising MCPSs in such datasets has been presented in~\cite{MartinSalvador2016a}, though the adaptation mechanisms for SMBO methods require further systematic research and form one of our future work directions.

In addition, it would also be valuable to investigate if using different data partitioning like Density-Preserving Sampling (DPS~\cite{Budka2010c}) would make any difference in the optimisation process. We believe that DPS could have a considerable impact when using certain datasets in SMAC strategy since SMAC discards potential poor solutions early in the optimisation process based on performance on only a few CV folds. In case the folds used are not representative of the overall data distribution, which as shown in~\cite{Budka2013} can happen quite often with CV, the effect on the solutions found can be detrimental.

Finally, at the moment, available SMBO methods only support single objective optimisation. However, it would be useful to find solutions that optimise more than one objective, including for instance a combination of prediction error, model complexity and running time as discussed in~\cite{Al-Jubouri2014}.

\bibliographystyle{IEEEtran}
\bibliography{IEEEabrv,references}
\vspace*{-1.5cm}
\begin{IEEEbiography}[{\includegraphics[width=1in,height=1.25in,clip,keepaspectratio]{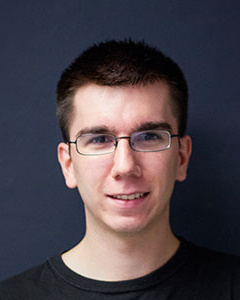}}]{Manuel Martin Salvador} graduated with distinction as a Computer Engineer from the University of Granada (Spain) in 2009, where he also received the MSc in Soft Computing and Intelligent Systems in 2011. Since 2017 he holds a PhD degree in Machine Learning from Bournemouth University (UK). Manuel has worked in a range of R\&D projects for the industry in several sectors: renewable energy, process industry and public transport. 
His main research areas include automatic data preprocessing, predictive modelling and adaptive systems.
\end{IEEEbiography}
\vspace*{-1.5cm}
\begin{IEEEbiography}[{\includegraphics[width=1in,height=1.25in,clip,keepaspectratio]{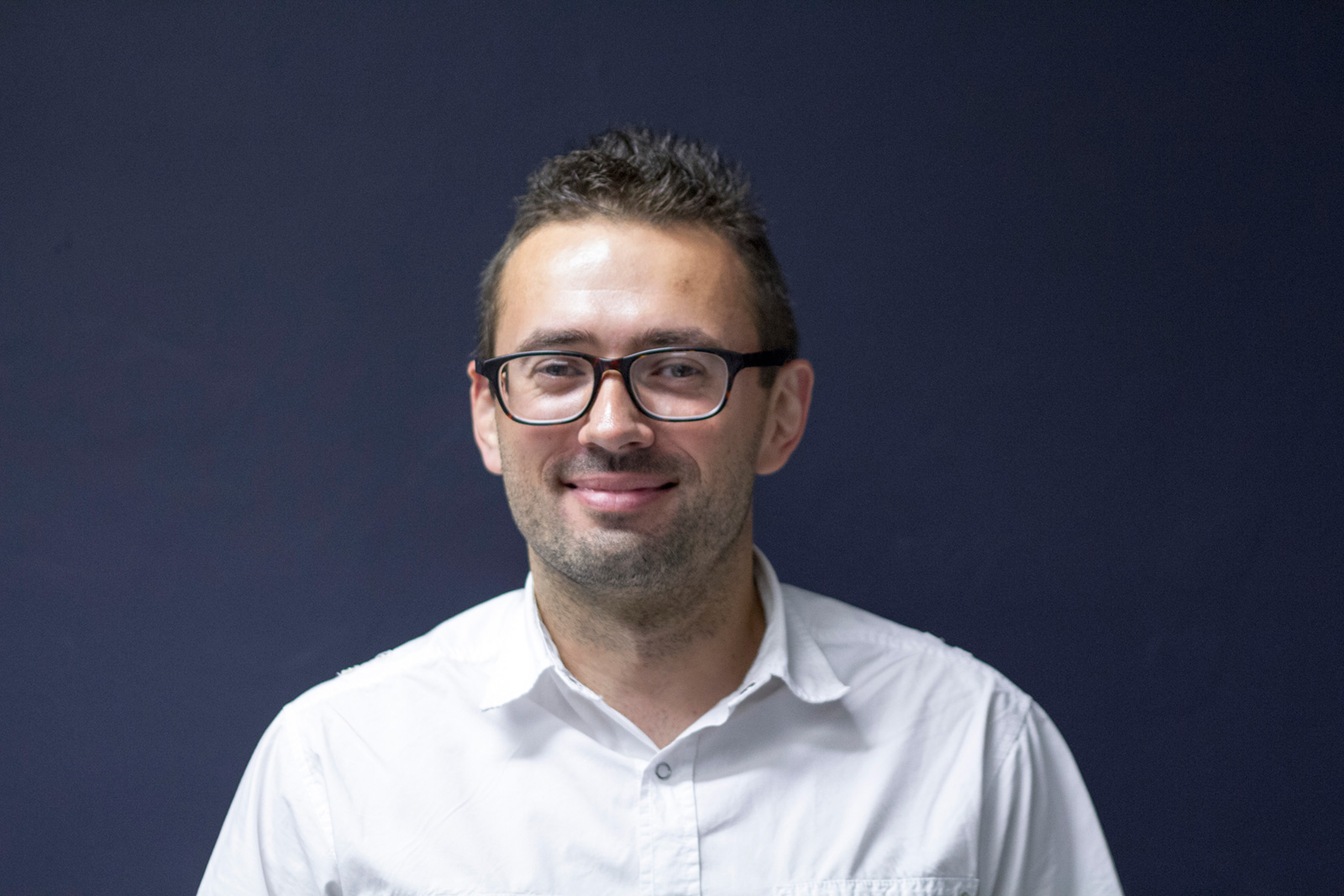}}]{Marcin Budka} is a Professor of Data Science in the Faculty of Science and Technology at Bournemouth University, UK. He received his dual MSc/BSc degree in Finance and Banking from the Katowice University of Economics (Poland, 2003), BSc in Computer Science from the University of Silesia (Poland, 2005) and PhD in Computational Intelligence from Bournemouth University (UK, 2010). 
His research interests lie in a broadly understood area of machine learning and data science, with a particular focus on practical applications. 
\end{IEEEbiography}
\vspace*{-1.5cm}
\begin{IEEEbiography}[{\includegraphics[width=1in,height=1.25in,clip,keepaspectratio]{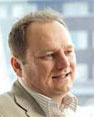}}]{Bogdan Gabrys} is a Data Scientist and a Professor of Data Science at the Advanced Analytics Institute, University of Technology Sydney, Australia. He received an MSc degree in Electronics and Telecommunication (Specialization: Computer Control Systems) from the Silesian Technical University, Poland in 1994 and a PhD in Computer Science from the Nottingham Trent University, UK in 1998. Over the last 20 years, Prof. Gabrys has been working at various universities and R\&D departments of commercial institutions. 
\end{IEEEbiography}

\end{document}